\documentclass[10pt,journal,compsoc]{IEEEtran}

\usepackage{epsfig}
\usepackage{graphicx}
\usepackage{amsmath}
\usepackage{amssymb}
\usepackage{subfigure}
\usepackage{graphicx}
\usepackage{caption}
\usepackage{longtable}
\usepackage{multirow}
\usepackage{algorithm}
\usepackage{algorithmicx}
\usepackage{algpseudocode}
\usepackage{amsmath}
\usepackage{indentfirst} 
\usepackage{verbatim}
\usepackage{color} 
\usepackage{verbatim}
\usepackage{threeparttable}
\usepackage{tablefootnote}
\usepackage{diagbox}
\usepackage{booktabs}
\usepackage[switch]{lineno}

\begin{document}
	
	\title{Variational Distillation for Multi-View Learning}

	\author{Xudong Tian,
		Zhizhong Zhang,
		Cong Wang,
		Wensheng Zhang,
		Yanyun Qu,
		Lizhuang Ma,\\
		Zongze Wu,
		Yuan Xie,~\IEEEmembership{~Member,~IEEE},
		Dacheng Tao,~\IEEEmembership{~Fellow,~IEEE}
		\IEEEcompsocitemizethanks{
			\IEEEcompsocthanksitem X. Tian, Z. Zhang, and Y. Xie are with School of Computer Science and Technology, East China Normal University, Shanghai, 200062, China; E-mail: \{52215901004, zzzhang, yxie\}@cs.ecnu.edu.cn
			\IEEEcompsocthanksitem C. Wang is with the Distributed and Parallel Software Laboratory, 2012 Labs, Huawei Technologies, Hangzhou, China; E-mail: wangcong64@huawei.com
			\IEEEcompsocthanksitem W. Zhang is with Institute of Automation, Chinese Academy of Sciences, Beijing, 100190, China; E-mail: wensheng.zhang@ia.ac.cn
			\IEEEcompsocthanksitem Y. Qu is with School of Information Science and Technology, Xiamen University, Fujian, 361005, China; E-mail: yyqu@xmu.edu.cn
			\IEEEcompsocthanksitem L. Ma is with School of Computer Science and Techology, East China Normal University, Shanghai, and also with the School of Electronic Information and Electrical Engineering, Shanghai Jiao Tong University, China; E-mail: lzma@cs.ecnu.edu.cn
			\IEEEcompsocthanksitem Z. Wu is with College of Mechatronics and Control Engineering, Shenzhen University, Shenzhen, China; E-mail: zzwu@szu.edu.cn
			\IEEEcompsocthanksitem D. Tao is with JD Exploer Academy, China and the University of Sydney, Australia; E-mail: dacheng.tao@gmail.com}
\thanks{}}

\IEEEtitleabstractindextext{%
\begin{abstract}

Information Bottleneck (IB) based multi-view learning provides an information theoretic principle for seeking shared information contained in heterogeneous data descriptions. However, its great success is generally attributed to estimate the multivariate mutual information which is intractable when the network becomes complicated. Moreover, the representation learning tradeoff, {\it i.e.}, prediction-compression and sufficiency-consistency tradeoff, makes the IB hard to satisfy both requirements simultaneously. In this paper, we design several variational information bottlenecks to exploit two key characteristics ({\it i.e.}, sufficiency and consistency) for multi-view representation learning. Specifically, we propose a Multi-View Variational Distillation (MV$^2$D) strategy to provide a scalable, flexible and analytical solution to fitting MI by giving arbitrary input of viewpoints but without explicitly estimating it. Under rigorously theoretical guarantee, our approach enables IB to grasp the intrinsic correlation between observations and semantic labels, producing predictive and compact representations naturally. Also, our information-theoretic constraint can effectively neutralize the sensitivity to heterogeneous data by eliminating both task-irrelevant and view-specific information, preventing both tradeoffs in multiple view cases. To verify our theoretically grounded strategies, we apply our approaches to various benchmarks under three different applications. Extensive experiments to quantitatively and qualitatively demonstrate the effectiveness of our approach against state-of-the-art methods.

\end{abstract}

}

\maketitle

\IEEEdisplaynontitleabstractindextext

%
\IEEEpeerreviewmaketitle


\section{Introduction}\label{introduction}
\IEEEPARstart{A}s more and more real-world data are collected from diverse sources or obtained from different feature extractors, multi-view representation learning has gained increasing attention due to its strong predictive power. For example, in auto-driving scene, there usually exist camera sensors that assist LiDAR data to perceive the complex 3D visual world, and therefore it enables us to take advantages of depth, texture and color information provided by multiple sensors to improve the predictive performance. From this perspective, multi-view learning aims to integrate various features ({\it i.e.}, heterogeneous data or visual descriptors) of the same object to promote the performance of existing machine learning system.


To effectively explore multi-view data, many efforts have been devoted to learning a consistent representation for discriminative information mining, such as Canonical Correlation Analysis (CCA) \cite{DCCA,DCCAE} or feature alignment \cite{alignment-representation}, both of which primarily maximize the similarity between representations from different viewpoints, and are more prone to introduce non-predictive redundancy and even cause considerable loss of predictive information.


Among various solutions, information bottleneck (IB) provides an information-theoretic principle \cite{ib} for multi-view learning, which has been successfully applied to a wide range of applications. \cite{mib,MVIB,CMIB-Nets}.
The central role of IB is to fit mutual information (MI) to maximize the correlation between representation and predictive information, while avoiding encoding task-irrelevant information. However, practical use of IB remains a persistent challenge due to the notorious difficulty of estimating mutual information. To deal with this, a common practice is to adopt the trainable parametric neural estimators \cite{infomax,mib,variationalbound} involving reparameterization trick, sampling, estimation of posterior distribution \cite{mine}, which, unfortunately, have relatively poor scalability in practice, and even become intractable when the network is complicated.



Another principal drawback of the information bottleneck is that, its optimization objective is essentially a trade-off between having a concise representation and achieving good predictive power, which makes it impossible to realize both high compression and accurate prediction \cite{ib,vib,dualIB}. Worse still, when dealing with heterogeneous data descriptions \cite{mib,MVIB,CMIB-Nets}, it has to struggle with multivariate mutual information to strike a balance between complementarity and consistency. Therefore, we raise a critical question: How to effectively explore the useful predictive information to learn a meaningful representation from multi-view data?




In this paper, we propose a new multi-view information bottleneck strategy, named as Multi-View Variational Distillation (MV$^2$D), for generalized multi-view representation learning. In this framework, we use variational inference to reconstruct the objective of IB and provide an analytical solution to MV$^2$D, which drives the network towards learning concise yet predictive representations under a consistent training goal, by \textbf{fitting mutual information without explicitly estimating it.} Specifically, MV$^2$D consists of two components ({\it i.e.}, sufficiency and consistency), where the first one enables us to preserve sufficient task-relevant information, while simultaneously discarding task-irrelevant distractors; and the second one neutralizes the sensitivity to multi-view data by refining the consistent information.

MV$^2$D is applicable to arbitrary input of viewpoints and can identify the prioritization for each representation by automatically exploring the useful consistent multi-view information. 
The resulting representations are then improved with enhanced generalization ability and robustness to the heterogeneous gap among different viewpoints. In addition, we show existing cutting-edge variational information bottleneck like Variational Self-Distillation (VSD), Variational Cross Distillation (VCD), Variational Mutual Distillation (VMD) \cite{VSD} are the special cases of our MV$^2$D framework. MV$^2$D and all its variants do not require any strong assumptions or mutual information estimators, and can concurrently attain two key characteristics of representation learning ({\it i.e.,} sufficiency and consistency) under rigorously theoretical guarantee.

To verify our theoretically grounded strategies in singe-view, cross-view and multi-view cases, we apply our approaches to the tasks of: (i) Cross-modal person re-identification\footnote{We do not explicitly distinguish multi-view and multi-modal throughout this paper.}; (ii) Multi-view classification; (iii) LiDAR-RGB semantic segmentation. Extensive experiments conducted on the widely adopted benchmark datasets demonstrate the effectiveness, robustness and superior performance of our approaches against state-of-the-arts methods. Our main contributions are summarized as follows:
\begin{itemize}
\item We design a new information bottleneck strategy for multi-view representation learning, with arbitrary input of views, named as Multi-View Variational Distillation (MV$^2$D), which is able to prevent both the prediction-compression and sufficiency-consistency trade-off, leading to a predictive and consistent representation.

\item Under strictly mathematical proofs, we introduce a generalized analytical solution to maximizing consistent information among multiple heterogeneous data observations, which significantly improves the robustness to view-changes by accurately eliminating both the view-specific and non-predictive details.

\item We show the proposed variational distillation framework could be flexibly applied to diverse multi-view tasks. The experiments on Cross-modal person re-identification; Multi-view classification and LiDAR-RGB semantic segmentation demonstrate the effectiveness of our approach.

\end{itemize}

\section{Related Work}\label{related work}
\subsection{Information Bottleneck}
The seminal work of Information Bottleneck is from \cite{ib,good_representation}, which introduces the general idea of using the information theoretic objective to train a deep model. But, unfortunately, they did not include any experimental results, since the optimization for IB relied on the iterative Blahut Arimoto algorithm, which is infeasible over high-dimensional variables, {\it e.g.,} deep neural networks (DNNs) \cite{vib}. On this basis, a series of explorations from theoretic study to practical use of IB principle have been witnessed.

{\textbf{Theoretic Study of IB.}} By using the variational inference and re-parameterization tricks, VIB \cite{vib} constructs a lower bound on IB objective (Eq. (\ref{objective of ib})), and enables the DNNs to handle the high-dimensional and continuous data under the guidance of IB. It avoids the restrictions that data must follow discrete or Gaussian distribution. Moreover, with the introduction of dual distortion \cite{dual_distortion}, dualIB framework is presented in \cite{dualIB}, which shifts the research attention from training to prediction, leading to the better stability compared with the classic IB structure. However, these methods still suffer from IB's
disadvantage ({\it i.e.}, the trade-off between prediction and compression) and consequently have unsatisfactory applicability for real world problems.

{\textbf{Practical Applications of IB.}} Typically, \cite{vdb} applies Eq. (\ref{objective of ib}) to the generative adversarial networks (GANs), with small modifications for a more robust generation process. \cite{IB_fine-tuning} alleviates the overfitting of large-scale pretrained language models in low-resource scenarios by directly introducing Eq. (\ref{objective of ib}). Other applications involving IB include decision-making system \cite{IBA}, speech processing \cite{speechprocessing}, ensemble learning \cite{DIBS}, neuroscience \cite{neuroscience} and deep neural networks understanding \cite{understandDNNs_4}. However, they either heavily rely on the mutual information estimator, or reformulate the IB objective based on strong assumptions, resulting in inferior practicality.

In contrast to all of the above, our work is the first to provide an analytical solution to fitting the mutual information without estimating it. The proposed VSD can better preserve task-relevant information, while simultaneously getting rid of redundant nuisances.

\subsection{Representation Learning}
The performance of machine learning methods is heavily dependent on the learned representations, which may entangle or hide different explanatory factors behind the data. Hence, a great deal of researches is devoted to designing data processing pipelines or transformations to attain representations that can support effective machine learning. Specifically, early works prefer feature engineering (we refer readers to \cite{survey_of_representation} for comprehensive studies) to take advantage of human ingenuity and prior knowledge. After that, numerous deep models adopt reconstruction-based \cite{DCCAE} representation, which enforces similarity constraint between the input and reconstructed output; or contrastive methods \cite{contrastive_learning}, which learn representations by maximizing similarity between augmentations of the same data point and minimizing similarity between different data points.

Unfortunately, feature engineering is labor-intensive, and shows inferiority to extract discriminative information in complex circumstances \cite{survey_of_representation}. Moreover, reconstruction-based and contrastive methods highly rely on similarity maximization, which are more prone to overfitting or obtain a trivial solution when encountering small scale datasets. On the contrary, our approach enables the model to accurately preserve predictive information {\it w.r.t.} the given task while discarding those superfluous under the information theoretic constraint towards generalized
representation learning. More importantly, our approach is quite efficient since it does not require large-scale training data or batch size. 

\begin{table}[t]	
\centering
\renewcommand{\arraystretch}{1.25}
\scriptsize
\caption{Basic notations and their descriptions.}\label{notations}
\begin{tabular}{|l|l|}
\hline
Notation & Meaning \\ \hline 
$x$, $y$& object, and the corresponding ground-truth label \\
$v$, $z$& observation, representation of $x$\\\hline
$\{x_1,...,x_n\}$ & multiple views of the same object $x$ \\
$\{v_1,...,v_n\}$ & observations collected from different viewpoints\\
$\{z_1,...,z_n\}$ & multi-view representations, denoted as $z_{\{1,...,n\}}$\\
$z_{\{1,...,n\}/i}$ & the entire $\{z_1,...,z_n\}$ but excluding $z_i$\\\hline
$I(v;z)$& mutual information, denoted as MI for simplicity \\
$I(v;z|y)$& conditional MI, abbreviated as $I_{vz|y}$ in diagrams\\
$I_{z_1|yz_2z_3}$&information unique to $z_1$, {\it i.e.}, $H(z_1|y,z_2,z_3)$\\
$I_{yz_1|z_2z_3}$&conditional MI between $y$ and $z_1$, {\it i.e.}, $I(y;z_1|z_2,z_3)$\\
$I_{yz_1z_2|z_3}$&MI among $y$,$z_1$ and $z_2$, {\it i.e.}, $I(y;z_1;z_2|z_3)$\\
$I_{yz_1z_2z_3}$ &information shared by $y$, $z_1$, $z_2$, $z_3$, {\it i.e.}, $I(y;z_1;z_2;z_3)$\\\hline
$E_{\theta}$, $E_{\phi}$ & the encoder and information bottleneck\\
$H(\cdot)$, $H(\cdot|\cdot)$& Shannon entropy, conditional entropy\\
$\mathbb{P}_{v}$, $\mathbb{P}_{z}$& predictions conditioned on $v$ and $z$ \\
$\mathbb{P}_{z_{\{1,...,n\}/i}}$ & prediction conditioned on $z_{\{1,...,n\}/i}$\\ \hline
\end{tabular}
\vspace{-0.9mm}
\vspace{-1.6mm}
\vspace{-0.9mm}
\end{table}

\subsection{Multi-Modal/View Representation Learning}
Multi-modal representation learning aims to build models that can process and relate information from multiple modalities, and its main difficulty is to explicitly measure the content similarity between the heterogeneous samples. Classical and deep multi-view representation learning methods can be roughly divided into joint representation \cite{joint-representation,DCCA}, alignment representation \cite{alignment-representation,CMAlign}, as well as shared and specific representation \cite{MvDA,MvNNcor}. The key idea of these methods is to establish a common representation space by exploring the semantic relationship among multi-view data. Please refer to \cite{survey_of_multi_view,survey_of_multi_view_2} for a comprehensive review.


Recently, another line of works ({\it e.g.}, MIB \cite{mib} and MVIB \cite{MVIB}) extend IB principle to the multi-view representation learning, which achieves promising results. For example, MIB integrates the heterogeneous representations by introducing a variational bound of Eq. (\ref{objective of ib}). However, it still requires explicit estimation to the mutual information, and is applicable when only two views are entailed. Consequently, it shows weak scalability for complex cases, and fails to learn a generalized representation for arbitrary input of viewpoints. By comparison, the proposed MV$^2$D avoids both the prediction-compression trade-off and sufficiency-consistency trade-off, enabling us to learn a predictive yet compact representation. The empirical study also demonstrates that our method can effectively eradicate both the non-predictive and view-specific information and thus significantly improve the robustness and generalization ability.

\subsection{Knowledge Distillation}
Knowledge distillation (KD) is a representative technique utilized for model compression and acceleration, which typically intends to learn a small student model from a large-scale teacher by minimizing a KL-divergence loss between their predictions. In addition to the student-teacher paradigm, recent developments have been extended to assistant-learning \cite{assistant-learning}, mutual-learning \cite{DML} and self-learning \cite{self-distillation} (please refer to \cite{survey_of_KD} for more related works). Despite the wide use of KD, fundamental analysis of what information should be distilled is still lacking in the literature, remaining the mechanism of KD still unclear.

\begin{figure}[t]
\centering
\renewcommand{\figurename}{Figure}
\includegraphics[width=1\linewidth]{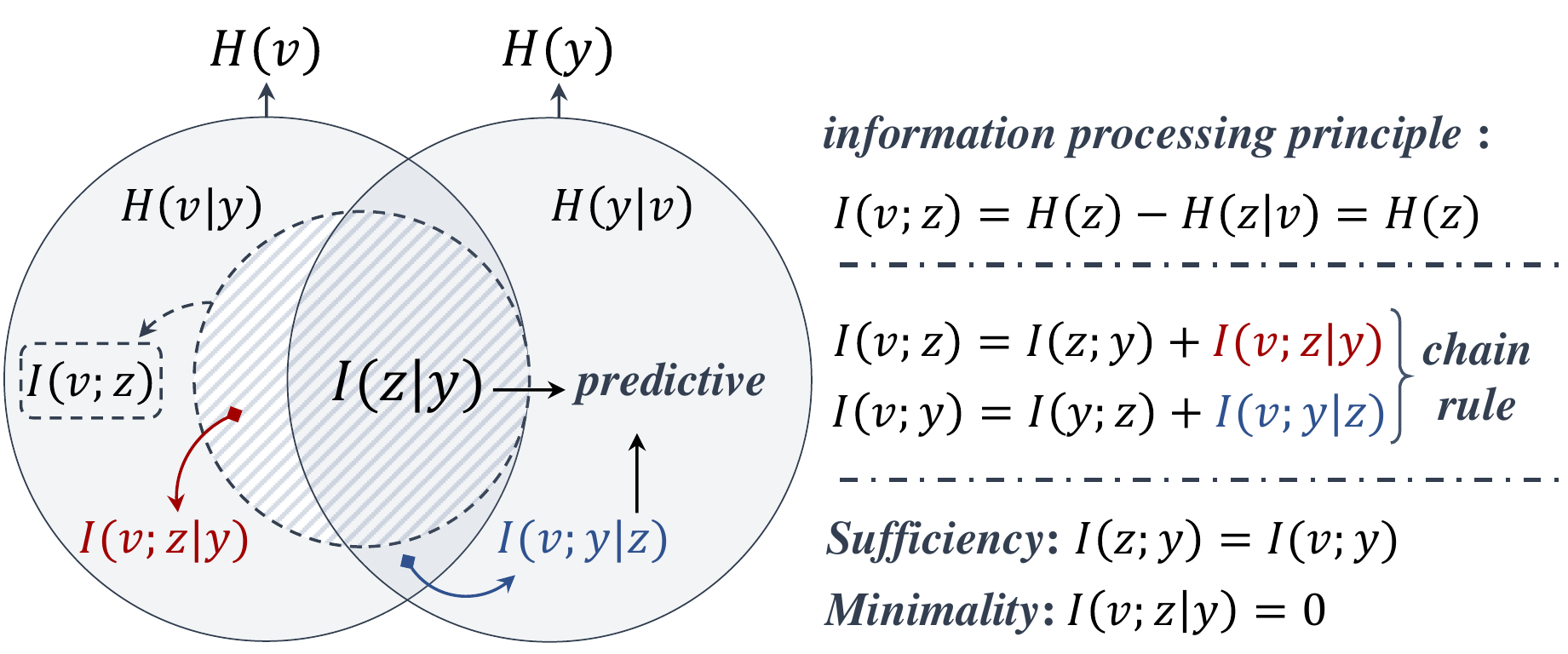}
\caption{Venn diagram visualization of entropies and mutual information for three variables $v$, $z$ and $y$.}
\label{Venn_VSD}
\vspace{-0.9mm}
\vspace{-1.6mm}
\end{figure}

\section{Preliminary}
In this section, we first provide a brief review of the IB principle \cite{ib} in the context of supervised learning. Then we introduce several variations of mutual information encompassed in our method. We summarize the basic notations and their descriptions in Tab. \ref{notations}.

\subsection{Information Bottleneck Principle}\label{IBP}
Given data observations $V$ and labels $Y$, the goal of representation learning is to obtain an encoding $Z$ which is maximally informative {\it w.r.t} $Y$ ({\it i.e.}, \textbf{sufficiency} in Fig. \ref{Venn_VSD}), and without any additional information about $V$ ({\it i.e.}, \textbf{minimality} in Fig. \ref{Venn_VSD}), measured by mutual information, {\it i.e.,}
\begin{equation}
I(Z;Y)=\int p(z,y)\log\frac{p(z,y)}{p(z) p(y)}dzdy.
\end{equation} 
Based on the information processing principle, we illustrate the definitions of sufficiency and minimality in Fig. \ref{Venn_VSD}, where the areas of the three circles represent $H(v)$, $H(y)$ and $H(z)$. To encourage the encoding process to focus on the label information, IB was proposed to enforce an upper bound $I_c$ to the information flow from the observations $V$ to the encoding $Z$, by maximizing the following objective:
\begin{equation}\label{definition of ib}
\max I(Z;Y) ~ s.t.~I(Z;V)\le I_c.
\end{equation}
Eq. (\ref{definition of ib}) implies that a compressed representation can improve the generalization ability by ignoring irrelevant distractors in the original input.
By using a Lagrangian objective, IB allows the encoding $Z$ to be maximally expressive about $Y$ while being maximally compressive about $X$ by:
\begin{equation}\label{objective of ib}
\mathcal{L}_{IB}=I(Z;V)-\beta I(Z;Y),
\end{equation}
where $\beta$ is the Lagrange multiplier. However, it has been shown that it is impossible to achieve both objectives in Eq. (\ref{objective of ib}) practically \cite{mib,vib} due to the trade-off optimization between high compression and high mutual information. 

More significantly, estimating mutual information in high dimension imposes additional difficulties \cite{mine,variationalbound} for optimizing IB. As a consequence, it inevitably introduces irrelevant distractors and discards some predictive cues in the encoding process. In Sec. \ref{method}, we show how we design a new strategy to deal with these issues, and extend it to cross-modality learning, and even generalize to multi-view representation learning.

\subsection{The Chain Rule of Mutual Information}
The chain rule \cite{sufficiency,vib,mib} can be utilized to subdivide the mutual information into multiple terms ({\it e.g.}, $I(v;z)$ and $I(v;y)$ in Fig. \ref{Venn_VSD}), which are defined and visualized in Fig. \ref{Venn_VSD} and Fig. \ref{Venn_MVD}. On this basis, numerous variants measuring statistical dependencies among different variables can be expressed, such as conditional mutual information \cite{mib,sufficiency}, interaction information \cite{cross-domain_MI} and total correlation \cite{total_correlation}. 

However, in practice, estimating the mutual information with even the simplest form ({\it e.g.}, $I(v;z)$) in high dimension can be particularly challenging, let along other sophisticated variations. Thus, to deal this such issue and promote applicability of the IB principle, we next present an analytical solution to fitting the mutual information without explicitly estimating it in both single-view and multi-view cases.

\section{Method}\label{method}
Let $v\in V$ be an observation of input data $x \in X$ extracted from an encoder $E(v|x)$. The challenge of optimizing an information bottleneck can be formulated as finding an extra encoding $E(z|v)$ that preserves all label information contained in $v$, while simultaneously discarding task-irrelevant distractors. To this end, we show the key roles of two characteristics of $z$, ({\it i.e.}, \textbf{sufficiency} and \textbf{consistency}) based on the information theory, and design a multi-view variational information bottlenecks to keep both characteristics.

\subsection{Generalized Variational Distillation for Multi-View Representation Learning}\label{MVD}

\begin{figure}[t]
	\centering
	\renewcommand{\figurename}{Figure}
	\includegraphics[width=1.0\linewidth]{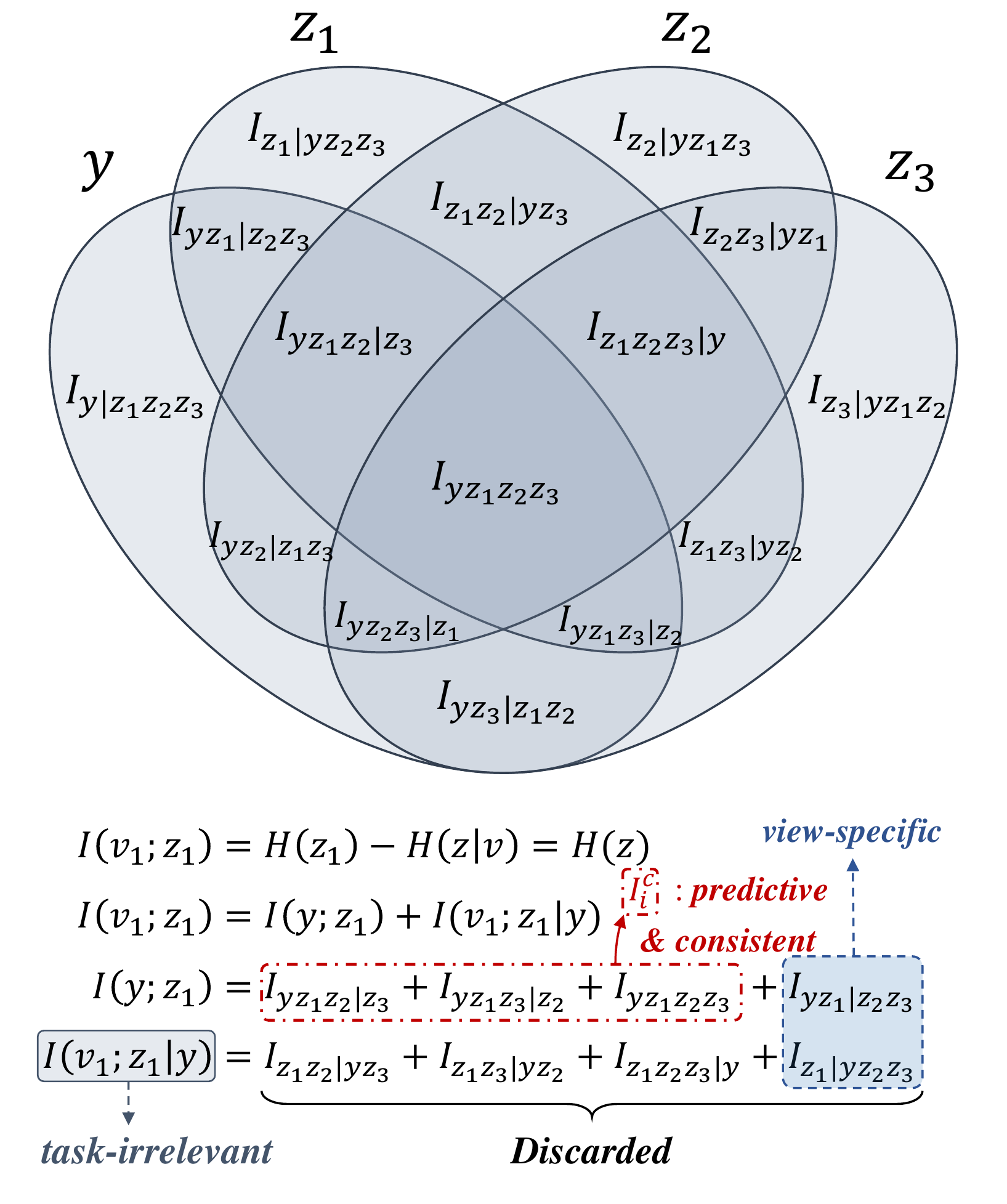}
	\vspace{-3.16mm}
	\caption{Venn diagram utilized to illustrate the mutual information among the target $y$ and $z_1$, $z_2$ and $z_3$. The areas of four ellipses denote the total information of each variable. Shaded areas denote the information shared by at least two variables. Some other information measures ({\it e.g.}, prediction and view-consistency) are indicated in the legend.}
	\label{Venn_MVD}
	\vspace{-0.9mm}
	\vspace{-1.6mm}
\end{figure}

Considering $\{v_1, v_2, ..., v_n\}$ are $n$ observations of $x$ that are collected from different viewpoints. An information bottleneck is used to produce representations $\{z_1, z_2, ..., z_n\}$ for keeping all predictive information {\it w.r.t} label $y$ while avoiding encoding task-irrelevant information. From this perspective, given a specific view $i$, the sufficiency of $z_i$ for $y$ could be defined as:
\begin{equation}
I(z_i;y)=I(v_i;y),\label{definition of sufficiency}
\end{equation}
where $v_i$ is the corresponding observation containing all label information. Previous work \cite{VSD} has shown that finding sufficiency representation {\it i.e.,} Eq. \ref{definition of sufficiency}, could be simplified to minimize the following objective:
\begin{equation}\label{obj1}
\min I(v_i;y)-I(z_i;y).
\end{equation}

This problem can be solved easily by the following theorem: 
~\\
\noindent{\textbf{Theorem 1. }}{\textit{Minimizing Eq. (\ref{obj1}) is equivalent to minimizing the subtraction of conditional entropy $H(y|z_i)$ and $H(y|v_i)$. That is:
\begin{flalign}
	&\min I(v_i;y)-I(z_i;y)\iff \min H(y|z_i)-H(y|v_i),\nonumber
\end{flalign} 
where $H(y|z_i):=-\int p(z_i)dz_i\int p(y|z_i)\log p(y|z_i)dy$.}} 
~\\

More specifically, given a sufficient observation $v_i$ for $y$, we have the following Corollary:

~\\
\noindent{\textbf{Corollary 1.} \textit{If the KL-divergence between the predicted distributions of a sufficient observation $v_i$ and the representation $z_i$ equals to $0$, then $z_i$ is sufficient for $y$ as well {\it i.e.},
\begin{equation}
	D_{KL}[\mathbb{P}_{v_i}||\mathbb{P}_{z_i}]=0 \implies H(y|z_i)-H(y|v_i)=0, \nonumber
\end{equation}
where $\mathbb P_z=p(y|z_i)$, $\mathbb P_v=p(y|v_i)$ represent the predicted distributions, and $D_{KL}$ denotes the KL-divergence.}}
~\\

The above theories reformulate the optimization objective of information bottleneck, which provides an analytical solution to achieving sufficiency for $\{z_1, z_2, ..., z_n\}$ {\it w.r.t.} the target $y$, separately. However, sufficiency is hard to attain because each representation can only partly describe the object in multi-view learning. Therefore, we introduce another strategy to promote consistency among the representations from different views.

In common practice \cite{survey_of_multi_view,survey_of_multi_view_2}, consistent information is simply defined as the consensus shared by different viewpoints, which, in practice, is usually indiscriminately required to learn. However, as illustrated in Fig. \ref{Venn_MVD}, consistent information is essentially composed of a series of terms when multiple viewpoints are involved. Thus there is a lack of guidance to seek the useful predictive information ({\it e.g.}, terms marked with red rectangle in Fig. \ref{Venn_MVD} for $z_1$, vice versa for $z_2$ and $z_2$) from multi-view data. In contrast to the conventional methods, our approach highlights the prioritization for different compositions.

Though information shared by all views usually leads to better generalization ({\it e.g.}, $I_{yz_1z_2z_3}$ in Fig. \ref{Venn_MVD}), it may not guarantee to be sufficient for the given task. In other words, representations learned by discarding all the information around the central area of Fig. \ref{Venn_MVD} ({\it i.e.}, $I_{yz_1z_2z_3}$) can hardly ensure predictive power for the downstream task {\it i.e.}, label $y$. Therefore, task-relevant information should be first preserved and then assigned with different weights according to the robustness to heterogeneous gaps among different views. For example, both $I_{yz_1z_2|z_3}$ and $I_{yz_1z_2z_3}$ in Fig. \ref{Venn_MVD} are supposed to be kept, and the later one should be given with larger weight since it is less sensitive to view-changes.

In the view of above, we define view-consistency to the generalized multi-view learning to clearly specify our goal. Formally, we have:

~\\{\textbf{Definition 2. Consistency:}} \textit{For any $z_i\in\{z_1,z_2,...,z_n\}$, it is view-consistent iff $I(v_i;z_i|y)+I(y;z_i|z_{\{1,...,n\}/i})=0.$}~\\

In particular, $I(v_i;z_i|y)$ denotes that the information contained in $z_i$ is unique to $v_i$ but is not predictive {\it w.r.t.} $y$, {\it i.e.}, superfluous information. $z_{\{1,...,n\}/i}$ represents the entire $\{z_1,z_2,...z_n\}$ but excluding $z_i$, thus $I(y;z_i|z_{\{1,...,n\}/i})$ is the information contained in $z_i$ but inaccessible to all other representations, {\it i.e.}, view-specific information. Intuitively, consistency requires elimination of both the task-irrelevant and view-specific distractors. To that end, we first factorize $I(v_i;z_i)$ using the chain rule \cite{mib}:

\begin{equation}
	I(v_i;z_i)= \underbrace{I(y;z_i)}_{\operatorname{predictive}}+\underbrace{I(v_i;z_i|y)}_{\operatorname{superfluous}}. \label{factorization5}
\end{equation}
Notice $I(y;z_i)$ is composed of multiple terms when two or more views are involved ({\it e.g.}, refer to the composition of $I(y;z_1)$ in Fig. \ref{Venn_MVD}). Thus, following the consistent principle in the multi-view learning, we further divide $I(y;z_i)$ as:
\begin{equation}
	I(y;z_i)=I(y;z_i|z_{\{1,...,n\}/i})+I^c_i,\label{factorization6}
\end{equation}
where we utilize $I^c_i$ to uniformly represent view-consistent information encoded within each $z_i$. For example, view-consistent information of $I^c_1$ {\it w.r.t} $z_1$ in Fig. {\ref{Venn_MVD}} is composed of $I_{yz_1z_2|z_3}$, $I_{yz_1z_3|z_2}$ and $I_{yz_1z_2z_3}$. This indicates $I^c_i$ essentially includes the predictive cues shared by both $z_i$ and all possible permutations of $\{z_1,z_2,...z_n\}/z_i$.


Based on the above analysis, an initial solution can be formulated by:
\begin{equation}
	\min \sum_{i \in n} \underbrace{-I^c_i }_{\operatorname{consistent}}+\underbrace{I(v_i;z_i|y)}_{\operatorname{superfluous}}+\underbrace{I(y ; z_{i} | z_{\{1 ,..., n\} /i})}_{\operatorname{view-spcific}}, \label{initial objective}
\end{equation}
which intuitively aims to eliminate both the task-irrelevant nuisances and view-specific information. Obviously, the min-max game of Eq. (\ref{initial objective}) is intractable to conduct. Thus, we present the following theory to equivalently reformulate our objective:

~\\
\noindent{\textbf{Theorem 3. }}{\textit{Given representations $\{z_1,...,z_n\}$ for $n$ different views, consistency can be promoted without violating the sufficiency constraint by:
		\begin{flalign}
			&\min \sum_{i \in n} D_{KL}\left[\mathbb{P}_{v_i}||\mathbb{P}_{z_i}\right]+D_{KL}[\mathbb{P}_{z_{\{1,...n\}}}||\mathbb{P}_{z_{\{1,...n\}/i}}],\nonumber
		\end{flalign}
		where $\mathbb{P}_{z_{\{1,...n\}}}=p(y|z_{\{1,...n\}})$, $\mathbb{P}_{z_{\{1,...n\}/i}}=p(y|z_{\{1,...n\}/i})$, and $\mathbb{P}_{v_i}=p(y|v_i)$ and $\mathbb{P}_{z_i}=p(y|z_i)$, all of them are essentially predicted distributions.
}} 
~\\

Detailed proof and formal assertions can be found in the Appendix. Hence, the refined training objective can be formed as:
\begin{equation}
	\begin{aligned}
		\mathcal{L}_{MV^2D}= \min _{\theta, \phi} \sum_{i\in n} \mathbb{E}_{v_i \sim E_{\theta}(v_i|x)}\mathbb{E}_{z_i \sim E_{\phi}(z_i|v_i)} \left[D_{KL}[\mathbb P_{v_i} \|\mathbb P_{z_i}]\right]\\
		+\sum_{i\in n} \mathbb{E}_{v_i \sim E_{\theta}(v_i|x)}\mathbb{E}_{z_i \sim E_{\phi}(z_i|v_i)} [D_{KL}[\mathbb{P}_{z_{\{1,...n\}}}||\mathbb{P}_{z_{\{1,...,n\}/i}}]]. \label{refined solution}
	\end{aligned}
\end{equation}
Here, $\theta$ and $\phi$ denote the parameters of the encoder and information bottleneck, respectively. In general, the first KL-divergence in Eq. (\ref{refined solution}) accounts for sufficiency, while the other one is utilized to promote consistency. Specifically, reducing $D_{KL}[\mathbb P_{v_i} \|\mathbb P_{z_i}]$ can eliminate the task-irrelevant nuisances and simultaneously maximize $I(y;z_i)$. On the other hand, minimizing $D_{KL}[\mathbb{P}_{z_{\{1,...n\}}}||\mathbb{P}_{z_{\{1,...,n\}/i}}]$ enables us to approximate $I^c_i$ to its upper bound ({\it i.e.}, $I(y;z_i)$) and thereby equivalently remove the view-specific information. 

\begin{figure}[t]
	\centering
	\renewcommand{\figurename}{Figure}
	\includegraphics[width=0.916\linewidth]{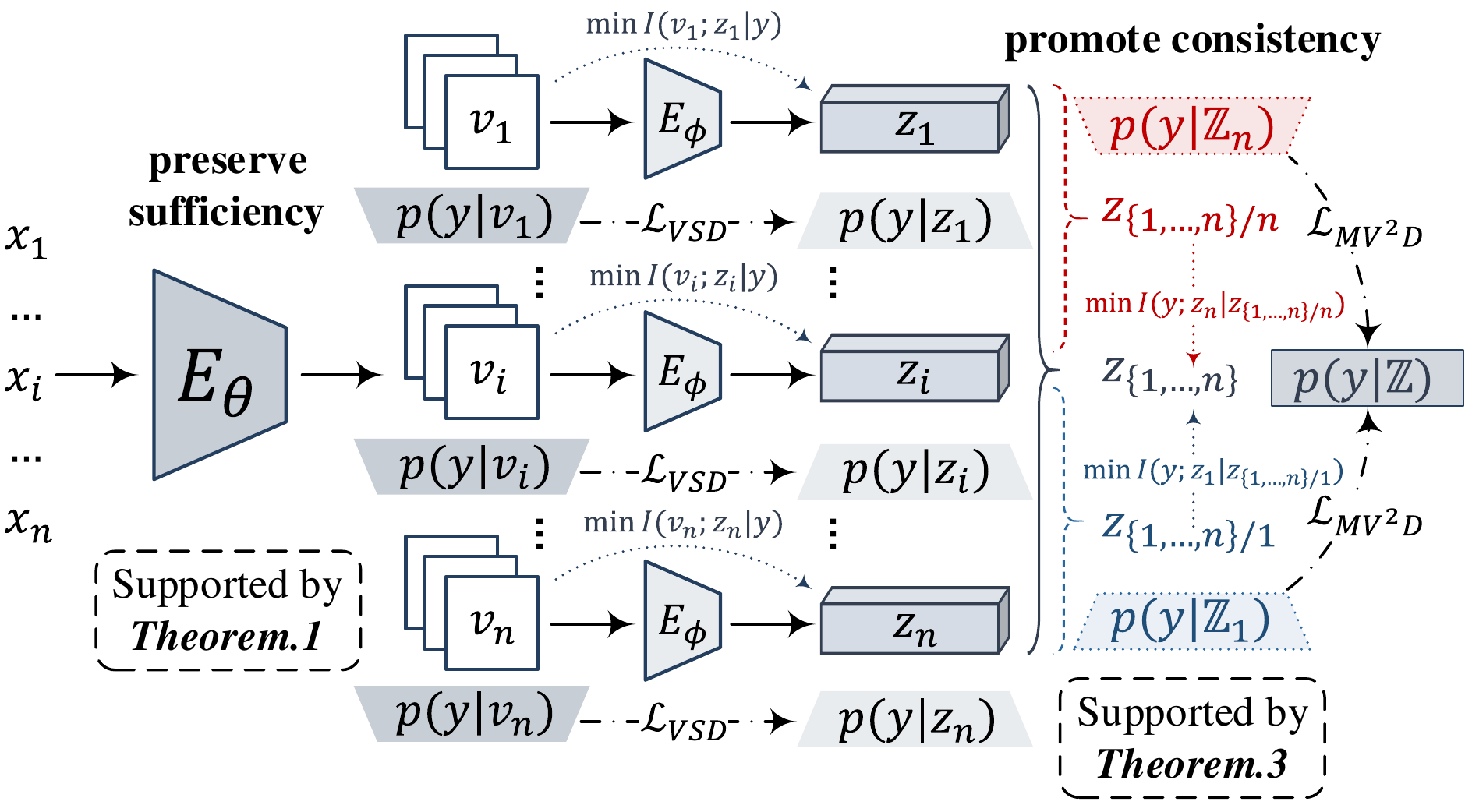}
	\caption{Illustration of the MV$^2$D framework, where $\mathbb{Z}$ refers to the whole $\{z_1,...,z_n\}$, while the subscription in $\mathbb{Z}_i$ denotes the index of the excluded variable ({\it e.g.}, $\mathbb{Z}_1$, $\mathbb{Z}_n$ exclude the first and the $n$-th representations, respectively).}
	\label{MVD_framework}
	\vspace{-0.9mm}
	\vspace{-1.6mm}
\end{figure}

We elaborate the resulting framework in Fig. \ref{MVD_framework}, where we demonstrate the process of eliminating both task-irrelevant and view-specific information. 

{\textbf{Discussion.}} Notice the MV$^2$D framework is essentially an extension of the information bottleneck architecture to the general multi-view learning, which reformulates the classic Lagrangian $I(V;Z)-I(Y;Z)$ as two terms of KL-divergence. Compared with other variants involving multiple views, domains, or modals learning \cite{mib,MVIB,cross-domain_MI,CMIB-Nets}, the primary advantages of MV$^2$D can be summarized as follows: (i) It ensures predictive and compact representations without computing any mutual information; (ii) It can accurately prioritize and purify the consistent information, which significantly improves the robustness to multi-view data. Proofs and detailed analysis of the advanced properties can be found in the Appendix.

\subsection{Example1 (Single-view): Variational Self-Distillation}\label{example1}
In this section, we present a special case of MV$^2$D to produce optimal representations within single viewpoint, which obtains an analytical solution to fitting the mutual information between an input $v$ and its representation $z$, namely Variational Self-Distillation (VSD) based on its formulation. Specifically, we first have $I(v;z)$ decomposed as:
\begin{equation}
I(v;z)=\underbrace{I(z;y)}_{\operatorname{predictive}}+\underbrace{I(v;z|y)}_{\operatorname{superfluous}}.\label{factorization1}
\end{equation}
As illustrated by Fig. \ref{Venn_VSD}, an optimal representation requires maximization of $I(z;y)$ ({\it i.e.}, sufficiency) and elimination of $I(v;z|y)$ ({\it i.e.}, minimality). To this end, we reformulate Eq. (\ref{factorization1}) based on the data processing inequality $I(z;y)\le I(v;y)$:
\begin{equation}
I(v;z)\le I(v;y)+I(v;z|y),\label{ineq1}
\end{equation}
which reforms the objective of an information bottleneck as: maximizing $I(v;y)$, minimizing $I(v;y)-I(z;y)$ and minimizing $I(v;z|y)$. Obviously, maximizing $I(v;y)$ is strictly consistent with the specific task and the last two terms are equivalent. Hence, the optimization is simplified to:
\begin{equation}
\min I(v;y)-I(z;y),
\end{equation}
which tackles the prediction-compression trade-off and can be equivalently achieved by Eq. (\ref{VSD}) based on Theorem 1 and Corollary 1. Formally, VSD maximizes the predictive information while concurrently minimizing the task-irrelevant nuisances through:
\begin{equation}
	\mathcal{L}_{VSD}=\min _{\theta, \phi} \mathbb{E}_{v \sim E_{\theta}(v|x)}\left[\mathbb{E}_{z \sim E_{\phi}(z|v)} \left[D_{K L}[\mathbb P_v \|\mathbb P_z]\right]\right],\label{VSD}
\end{equation}
where $\theta$ and $\phi$ denote the parameters of an encoder and information bottleneck, respectively. $\mathbb{P}_z=p(y|z)$, $\mathbb{P}_z=p(y|z)$ are the predictions, and $D_{KL}$ represents the KL-divergence.

Compared with other IB strategies \cite{ib,vib,vdb}, VSD simultaneously achieves both sufficiency and minimality without estimating mutual information, demonstrating superior scalability and flexibility. 

\subsection{Example2 (Cross-view): Variational Cross Distillation and Variational Mutual Distillation}\label{VCD and VMD}
To deal with typical cross-view issues ({\it e.g.}, LiDAR-RGB, infrared-visible), we introduce another two variations of the MV$^2$D, {\it i.e.}, Variational Cross Distillation (VCD) and Variational Mutual Distillation (VMD). Consider $v_1$ and $v_2$ as inputs from different modalities, and $z_1$, $z_2$ are the corresponding representations. VCD and VMD aim to neutralize the modal-discrepancies by eliminating both modal-specific and task-irrelevant information. More specifically, we first have the mutual information between $z_1$ and $v_1$ decomposed as follows (vice versa for $z_2$ and $v_2$):
\begin{equation}
	I(v_1;z_1)=I(v_1;z_1|v_2)+I(v_2;z_1),\label{factorization_VMD}
\end{equation}
\begin{equation}
	I(v_2;z_1)=I(v_2;z_1|y)+I(v_2;z_1;y).\label{factorization_VCD}
\end{equation}
As previously introduced, $I(v_1;z_1|v_2)$ represents that, the information contained in $z_1$ is unique to $v_1$ and is not accessible to $v_2$, {\it i.e.}, modal-specific information, and $I(z_1;v_2)$ denotes the information shared by $z_1$ and $v_2$, which is named as modal-consistent information. On the other hand, $I(v_2;z_1|y)$ denotes the irrelevant information encoded in $z_1$ regarding given task \cite{mib}, {\it i.e.}, superfluous information. Combining Eq. (\ref{factorization_VCD}) with Eq. (\ref{factorization_VMD}), we have:
\begin{equation}
	I(v_1;z_1)=\underbrace{I(v_1;z_1|v_2)}_{\operatorname{modal-specific}}+\underbrace{I(v_2;z_1|y)}_{\operatorname{superfluous}}+\underbrace{I(v_2;z_1;y)}_{\operatorname{predictive}}. \label{representation composition}
\end{equation}
On this basis, VMD and VCD are applied to eliminate $I(v_1;z_1|v_2)$ and $I(v_2;z_1|y)$ respectively through:
\begin{equation}
	\mathcal{L}_{VMD}=\min _{\theta, \phi}\mathbb{E}_{v_1,v_2 \sim E_{\theta}(v|x)}\mathbb{E}_{z_1,z_2 \sim E_{\phi}(z|v)} \left[D_{KL}[\mathbb P_{z_1} \|\mathbb P_{z_2}]\right], \label{VMD}
\end{equation}
\begin{equation}
	\mathcal{L}_{VCD}=\min _{\theta, \phi}\mathbb{E}_{v_1,v_2 \sim E_{\theta}(v|x)}\mathbb{E}_{z_1,z_2 \sim E_{\phi}(z|v)} \left[D_{KL}[\mathbb P_{v_2} \|\mathbb P_{z_1}]\right]. \label{VCD}
\end{equation}
Here, $\theta$ and $\phi$ also denote the parameters of an encoder and information bottleneck architecture, and $\mathbb{P}_{z_1}=p(y|z_1)$, $\mathbb{P}_{v_2}=p(y|v_2)$ are the predicted distributions. Note Eq. (\ref{VMD}) and Eq. (\ref{VCD}) are symmetrically conducted to $z_2$ and $v_2$ to eliminate $I(v_2;z_2|v_1)$ and $I(v_1;z_2|y)$ respectively.

In essence, both VCD and VMD are special cases of MV$^2$D when the total of viewpoints is two. It is noteworthy that they show remarkable scalability and flexibility when applying to various cross-modal issues with large-scale datasets, further demonstrating the effectiveness of MV$^2$D in diverse circumstances. Next, we reveal the connections between MV$^2$D and its variations.


\begin{figure}[t]
	\centering
	\includegraphics[width=1\linewidth]{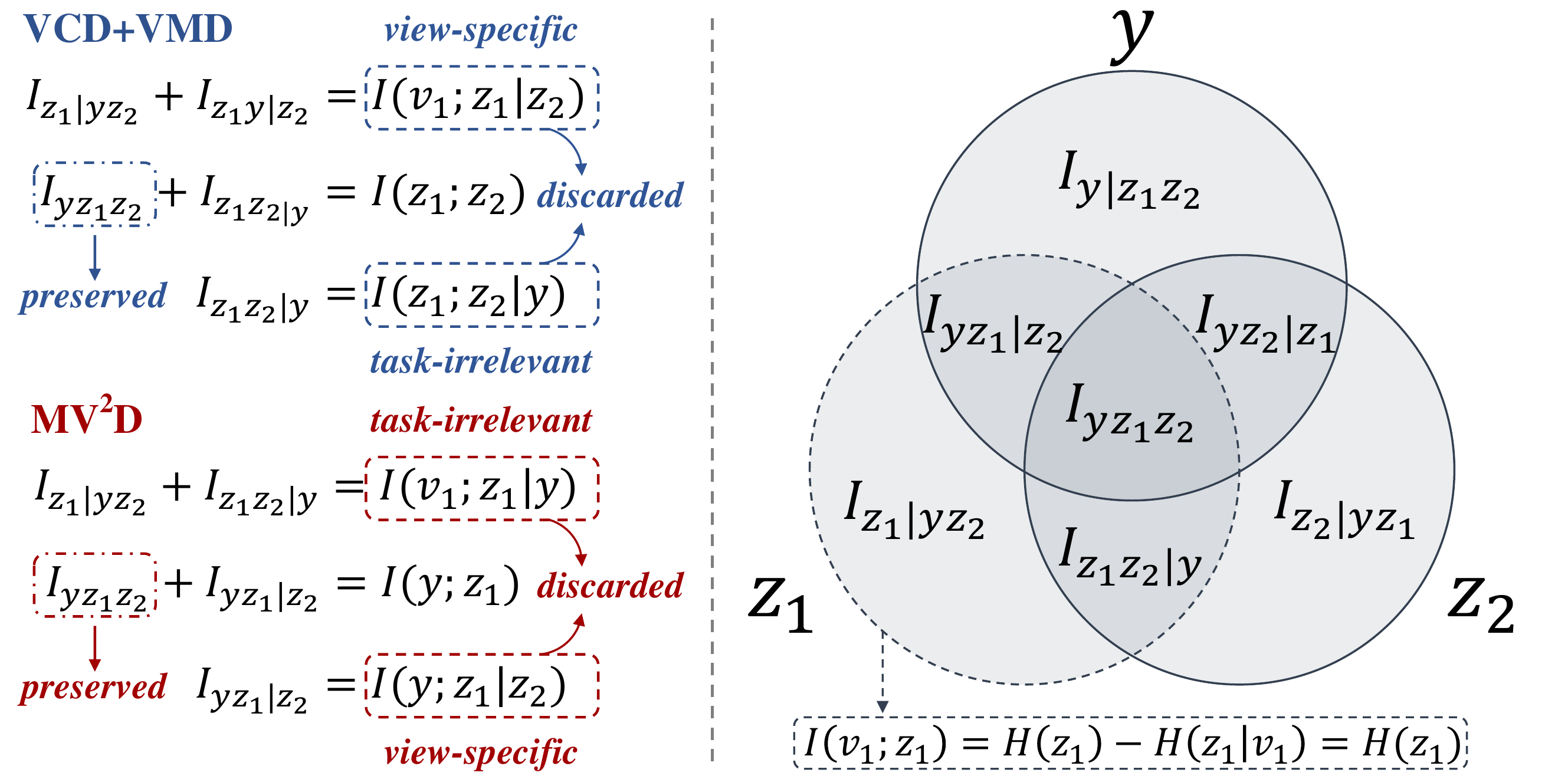}
	\caption{Venn diagram in the right half is utilized for describing the mutual information among two representations $z_1$, $z_2$ and the target $y$. Left part demonstrates the proof skeleton to Sec. \ref{Connections} from the view of mutual information.}\label{VCD_Venn}
\end{figure}

\subsection{Connections between Different Variations}\label{Connections}
Consider two special cases, where only one or two viewpoints are involved. For $n=1$, apparently $I(y;z)=I^c$, and Eq. (\ref{refined solution}) degenerates to VSD. For $n=2$, suppose $z_1$, $z_2$ are two representations corresponding to $v_1$ and $v_2$, both of which are two observations of the same object $x$ from two viewpoints. For better illustration, we first utilize the Venn diagram describing the mutual information among $z_1$, $z_2$, $y$ in Fig. \ref{VCD_Venn}, and give our proof as follows. 
	
	Recall the chain rule \cite{mib,sufficiency,conditional_MI}, we have:
	\begin{flalign}
		I(v_1;z_1)=&I(v_1;z_1|z_2)+I(z_2;z_1)\nonumber\\
		=&I(v_1;z_1|z_2)+I(z_2;z_1|y)+I^c_1. \label{MVD_10}
	\end{flalign}
	Revising the order of the chain, we have:
	\begin{flalign}
		I(v_1;z_1)=&I(v_1;z_1|y)+I(y;z_1)\nonumber\\
		=&I(v_1;z_1|y)+I(y;z_1|z_2)+\widehat{I}^c_1,
	\end{flalign}
	in which we use hatted symbols to distinguish consistent information preserved by VCD, VMD and MV$^2$D. Based on the definition of conditional mutual information, we have:
	\begin{flalign}
		I(v_1;z_1|z_2)&=H(z_1|z_2),\\
		I(z_2;z_1|y)&=H(z_1|y)-H(z_1|z_2,y).
	\end{flalign}
	Similarly, we have the following decomposition for Eq. (\ref{MVD_10}):
	\begin{flalign}
		I(v_1;z_1|y)&=H(z_1|y),\\
		I(y;z_1|z_2)&=H(z_1|z_2)-H(z_1|z_2,y). \label{MVD_11}
	\end{flalign}
	Note both $H(z_1|v_1,z_2)$ and $H(z_1|v_1,y)$ equals to zero based on the data processing inequality, thus they are omitted for simplicity. Combining Eq. (\ref{MVD_10})-Eq. (\ref{MVD_11}), we conclude that $I^c_1$ are equivalent to $\widehat{I}^c_1$, indicating Eq. (\ref{refined solution}) degenerates to Eq. (\ref{VCD}) and Eq. (\ref{VMD}) when the total of viewpoints is two.

In the view of above, we introduce the following theory to reveal the connections between MV$^2$D and its variations.

~\\
\noindent{\textbf{Corollary 2.} \textit{VSD, VCD and VMD are special cases of the proposed Multi-View Variational Distillation, in which there are only one or two views are involved.}}

\section{Applications}\label{applications}
In this section, we show the proposed variational distillation framework could be flexibly applied to various multi-modal/multi-view representation learning tasks: (i) Visible-Infrared Person Re-identification; (ii) Multi-view Classification; (iii) LiDAR-Image Semantic Segmentation. The quantitative and qualitative results demonstrate the effectiveness of our approach in various circumstances ({\it e.g.}, single-view, cross-view and multiple-view).

\begin{table*}[ht]
\centering
\renewcommand{\arraystretch}{1.25}
\footnotesize
\caption{Performance of the proposed method compared with state-of-the-arts. Note that all methods are measured by CMC and mAP on SYSU-MM01. {\color{blue}introduce 2-view and 3-view. }}\label{comparision1}
\begin{tabular}{l|l|c|c|c|c|c|c|c|c|c}
\hline
\multicolumn{3}{c|}{Settings} & \multicolumn{4}{c|}{All Search} & \multicolumn{4}{c}{Indoor Search}\\ \hline
Type &Method& Venue& Rank-1 & Rank-10 & Rank-20 & mAP & Rank-1 & Rank-10 & Rank-20 & mAP \\ \hline

Generative&Hi-CMD \cite{hi-cmd}& CVPR{\color{blue}'20} & 34.94 & 77.58 & - & 35.94& -& - & - & - \\ 
Network Design&DDAG \cite{ddag} & ECCV{\color{blue}'20} & 54.75& 90.39 & 95.81 & 53.02 & 61.02& 94.06 & 98.41 & 67.98 \\
Network Design&NFS\cite{NFS}& CVPR{\color{blue}'21} & 56.91 &91.34 &96.52 &55.45 & 62.79& 96.53 & 99.07 & 69.79 \\
Metric Design&MCLNet \cite{MCLNet} & CVPR{\color{blue}'21} & 65.40&93.33 &97.14 &61.98 & 72.56 & 96.98 & 99.20 & 76.58\\
Network Design&SMCL\cite{SMCL}& ICCV{\color{blue}'21} &67.39 &92.87 &96.76 &61.78&68.84& 96.55 & 98.77 &75.56\\
Network Design&CM-NAS \cite{CM-NAS} & ICCV{\color{blue}'21} & 61.99 &92.87 &97.25 &60.02 &67.01 &97.02 &99.32&72.95\\
Network Design&CMAlign \cite{CMAlign} & ICCV{\color{blue}'21} & 55.41 &- &- &54.14&58.46 & - & - &66.33\\
Network Design&AGW \cite{AGW} & TPAMI{\color{blue}'21} & 47.50 &84.39 &92.14 &47.65 & 54.17& 91.14 & 95.98 &62.97 \\
\hline
Representation&ours (baseline)& - &64.15 &94.42&98.68&61.74&69.61 & 95.78 & 98.90 &75.15\\
Representation&ours (2-view)& - &{\color{blue}\textbf{70.02}} & {\color{blue}\textbf{96.17}} &{\color{blue}\textbf{98.76}}& {\color{blue}\textbf{66.70}}& {\color{blue}\textbf{78.26}} & {\color{blue}\textbf{97.87}} & {\color{blue}\textbf{99.72}} &{\color{blue}\textbf{81.79}}\\
Representation&ours (3-view)& -& {\color{red}\textbf{71.65}} & {\color{red}\textbf{96.26}} &{\color{red}\textbf{98.71}}& {\color{red}\textbf{67.95}} & {\color{red}\textbf{79.08}} & {\color{red}\textbf{99.00}} &{\color{red} \textbf{99.91}}& {\color{red}\textbf{81.84}} \\ \hline
\end{tabular}
\end{table*}

\begin{table}[ht]
\centering
\renewcommand{\arraystretch}{1.25}
\footnotesize
\caption{Comparison with the state-of-the-arts on RegDB dataset under visible-thermal and thermal-visible settings.}\label{comparision2}
\begin{tabular}{l|c|c|c|c|c}
\hline
\multicolumn{2}{c|}{Settings} & \multicolumn{2}{c|}{Visible2Thermal} & \multicolumn{2}{c}{Thermal2Visible} \\ \hline
Method &Venue & Rank-1 & mAP& Rank-1 & mAP\\ \hline

Hi-CMD \cite{hi-cmd}&CVPR{\color{blue}'20} & 70.9 & 66.0 & -& -\\
DDAG \cite{ddag} &ECCV{\color{blue}'20}& 69.3 & 63.5 & 68.1 & 61.8 \\ 
NFS \cite{NFS}&CVPR{\color{blue}'21} & 80.5&72.1 &77.9 &69.8\\
MCLNet \cite{MCLNet}&CVPR{\color{blue}'21} & 80.3&73.1 &75.9&69.5\\
CMAlign \cite{CMAlign}&ICCV{\color{blue}'21} & 67.6&74.2&65.5 &65.9\\
AGW \cite{AGW}&TPAMI{\color{blue}'21} & 70.1 &66.4&70.5 &72.4\\
\hline
ours (baseline)&-& 79.9& 77.2& 77.5& 76.2 \\ 
ours (2-view)&-& {\color{blue}{\textbf{81.6}}}& {\color{blue}{\textbf{78.7}}}& {\color{blue}{\textbf{79.1}}} & {\color{blue}{\textbf{77.5}}} \\ 
ours (3-view)&-& {\color{red}{\textbf{83.1}}}& {\color{red}{\textbf{80.1}}}& {\color{red}{\textbf{81.2}}} & {\color{red}{\textbf{78.4}}} \\ 
\hline
\end{tabular}
\end{table}

\subsection{Cross-Modal Person Re-identification}
We first evaluate our approach on Visible-Infrared Person Re-identification task. In this application, there are two kinds of images from different modals ({\it i.e.}, infrared and visible), and the objective is to match the target person images among a gallery of images when given a query image from another modal. The key challenge of this task hence lies in the huge heterogeneous gap between the visible and infrared images, which requires both complementary and consistent information to facilitate cross-modal retrieval. To verify the effectiveness of MV$^2$D framework, we also design an intermediate modality by transforming both kinds of images to a new uniform image representation (see Fig. \ref{additional viewpoint}). 


\begin{figure}[t]
\centering
\renewcommand{\figurename}{Figure}
\includegraphics[width=1\linewidth]{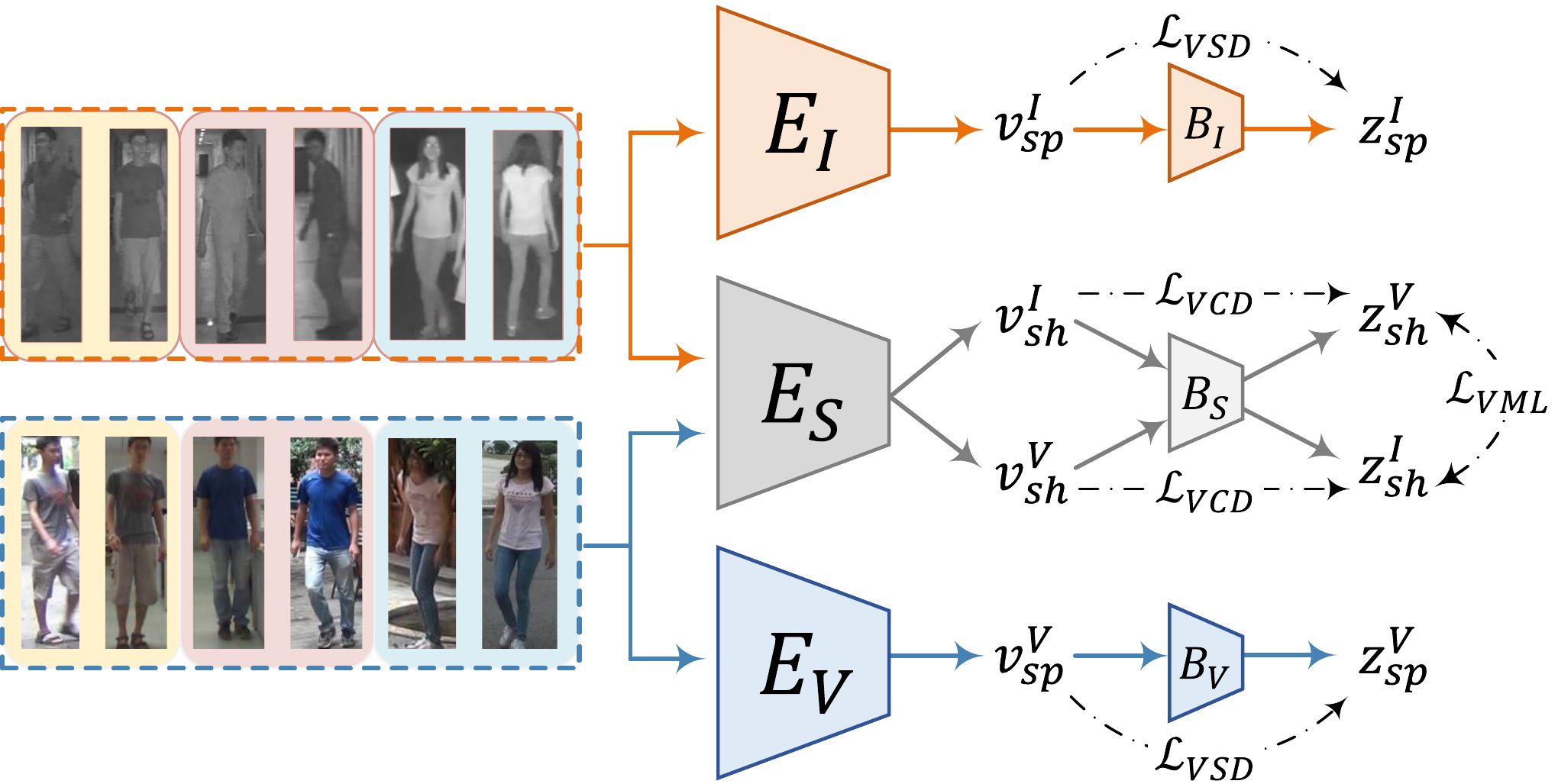}
\caption{Network architecture for Multi-Modal Re-ID. $E_{I/S/V}$ and $B_{I/S/V}$ represent the encoder (ResNet-50) and information bottleneck (multi-layer perceptrons), respectively. $v$ and $z$ denote the observations and representations from encoder and information bottleneck, respectively}
\label{reid framework}
\end{figure}

\subsubsection{Evaluation Protocol and Benchmarks}
In this section, we introduce the adopted benchmark datasets and corresponding evaluation standards. We follow the popular protocol \cite{zero-padding,AGW} for evaluation, where both cumulative match characteristic (CMC) and mean average precision (mAP) are used.

{\textbf{SYSU-MM01}} \cite{zero-padding} is collected from $6$ cameras of both indoor and outdoor environments. It contains $287,628$ visible images and $15,792$ infrared images of $491$ different persons in total, each of which is at least captured by two cameras. There are two search modes on SYSU-MM01, {\it i.e.}, all-search mode and indoor-search mode, and the difference lies in whether the outdoor cameras are excluded from the gallery. 

{\textbf{RegDB}} \cite{RegDB} is collected from two aligned cameras (one visible and one infrared) and it totally includes $412$ identities, where each identity has $10$ infrared images and $10$ visible images. Following the experimental protocol in \cite{RegDB}, we divide the dataset into training and test sets randomly, each of which includes non-overlapping 206 identities. We test our model in both visible-to-thermal and thermal-to-visible settings. 
The final reported results are averaged over $10$ trials with different training/test splits.

\subsubsection{Implementation Details}
{\textbf{Critical Architectures.}} For both MM01 and RegDB, we deploy three parallel branches, each of which is composed of a ResNet50 backbone ({\it i.e.}, encoder $E_{\theta}$) and an information bottleneck ({\it i.e.}, $E_{\phi}$: multi-layer perceptrons of $2$ hidden ReLU units of size $1,024$ and $512$ respectively with an output of size $2\times256$ that parameterizes mean and variance). In particular, we use two parallel modal-specific branches equipped with VSD to handle single modal image, and the remaining one ({\it i.e.}, modal-shared branch) takes cross-modal images as input, trained with VCD and VMD to produce consistent representations. For each branch, the backbone first encodes the input image to $2048$-D feature ({\it i.e.}, observation $v$), then it is forwarded to the information bottleneck to obtain the compressed representation $z$. See Fig. \ref{reid framework} for the illustration of our ReID framework.

\begin{figure}[t]
	\centering
\renewcommand{\figurename}{Figure}
\caption{Illustrative examples of the adopted viewpoints in 3-view experiments.}\label{additional viewpoint}
	\includegraphics[width=0.916\linewidth]{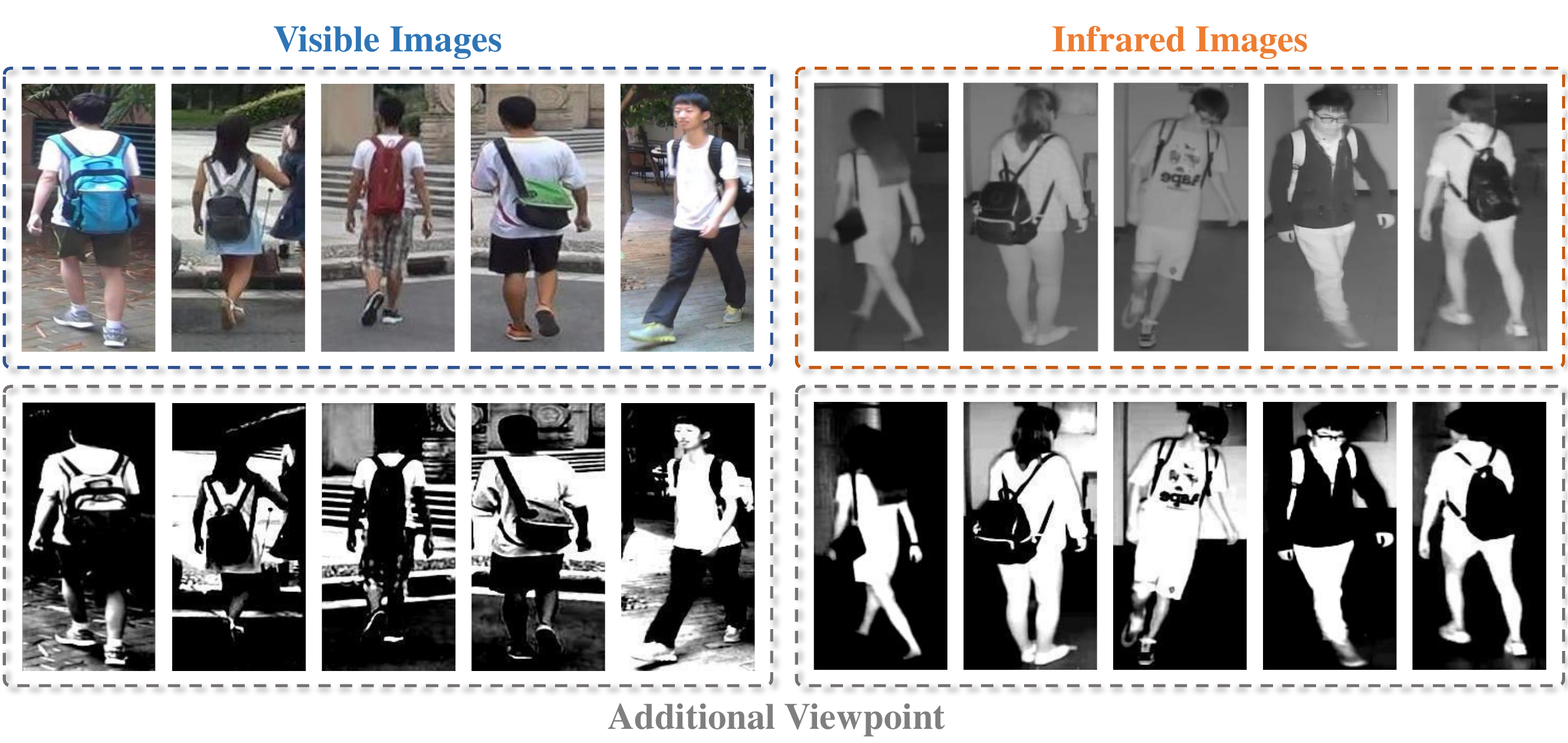}
	\vspace{-0.9mm}
	\vspace{-1.6mm}
	\end{figure}

Moreover, we design another uniform image representation in addition to the default visible and infrared ones (see Fig. \ref{additional viewpoint} for illustration). On this basis, the modal-shared branch takes inputs from three different viewpoints and adopts MV$^2$D to further investigate the effectiveness in 3-view circumstance (please refer to the supplementary materials for more details).

{\textbf{Training.}} Following \cite{SMCL,NFS,MCLNet}, we adopt the strong baseline with various training tricks, {\it i.e.}, warm up (linear scheme for first $10$ epochs) and label smooth. We utilize the rank loss \cite{rank_loss} for Re-ID learning, and set the weights of cross-entropy, rank loss and variational distillation objective to $1$, $1$, $2$, respectively. All experiments are optimized by Adam optimizer with an initial learning rate of $2.6\times10^{-4}$, which then decays $10$ times at $200$ epochs in total of $300$. Horizontal flip and normalization are utilized to augment the training images, where the images are resized to $288 \times144$. The batch size is set to $64$ for all experiments, in which it contains $8$ different identities, and each identity includes $4$ RGB images and $4$ IR images.

\subsubsection{Experimental Results}

{\textbf{Comparison.}} As shown in Tab. \ref{comparision1} and Tab. \ref{comparision2}, our approaches outperform all competitors by a large margin on both datasets. Moreover, compared with our baseline model, the proposed MV$^2$D significantly boosts the performance in both 2-view and 3-view cases, demonstrating its effectiveness and generalization ability. It is also noteworthy that our optimization is quite efficient since the estimation of mutual information is avoided (see complexity comparison in Tab. \ref{Complexity}). Since the additional viewpoint is not provided in the benchmarks by default, the following experiments are conducted only on the infrared and visible images.

{\textbf{Ablation Study.}} We first clarify different settings in Tab. \ref{effectiveness}, where ``$E_{S}$'' and ``$E_{I/V}$'' denote whether we use the modal-shared branch and modal-specific branches. ``$B_{S}$'' and ``$B_{I/V}$'' indicate that whether we utilize the information bottleneck architecture in each branch. ``CIB'' denotes we adopt the conventional IB for training. ``VSD'', ``VMD'' and ``VCD'' denote our approaches, and are uniformly represented with ``VD'' when applying all of them. Based on Tab. \ref{effectiveness}, we have the following observations: 

(i) Information bottleneck architecture can improve the performance in both singe-view, cross-view and triple-view cases (see $3^{\text{rd}}$, $7^{\text{th}}$ and $11^{\text{th}}$ row in Tab. \ref{effectiveness}).

(ii) It seems that conventional IB strategy has no advantages in learning predictive information (see $2^{\text{nd}}$, $6^{\text{th}}$ and $10^{\text{th}}$ row in Tab. \ref{effectiveness}). We conjecture it is because, by explicitly reducing $I(v;z)$, conventional IB may not recognize label information from task-irrelevant distractors, and probably discard all of them. On the other hand, estimation of mutual information in high dimension is difficult, especially when involving multi-modal data and latent variables in our setting, which leads to a sharp drop to performance.


(iii) Our approach provides remarkable improvement under various settings (see $4^{\text{th}}$, $8^{\text{th}}$ and $12^{\text{th}}$ row in Tab. \ref{effectiveness}). In single-view case, by maximally preserving predictive information while simultaneously reducing superfluous details, VSD outperforms the conventional IB by $27.68\%$@Rank-1 and $24.58\%$@mAP (comparing $8^{\text{th}}$ with $6^{\text{th}}$ row in Tab. \ref{effectiveness}). In cross-view case, VCD and VMD achieve $23.14\%$@Rank-1 and $22.82\%$@mAP improvement against the conventional IB (see $4^{\text{th}}$ and $2^{\text{nd}}$ row in Tab. \ref{effectiveness}), demonstrating huge advantages as well.

\begin{figure}[t]
\renewcommand{\figurename}{Figure}
\centering
\subfigure[$z_{sp}^V$ (VSD)]{
\includegraphics[width=0.2725\linewidth]{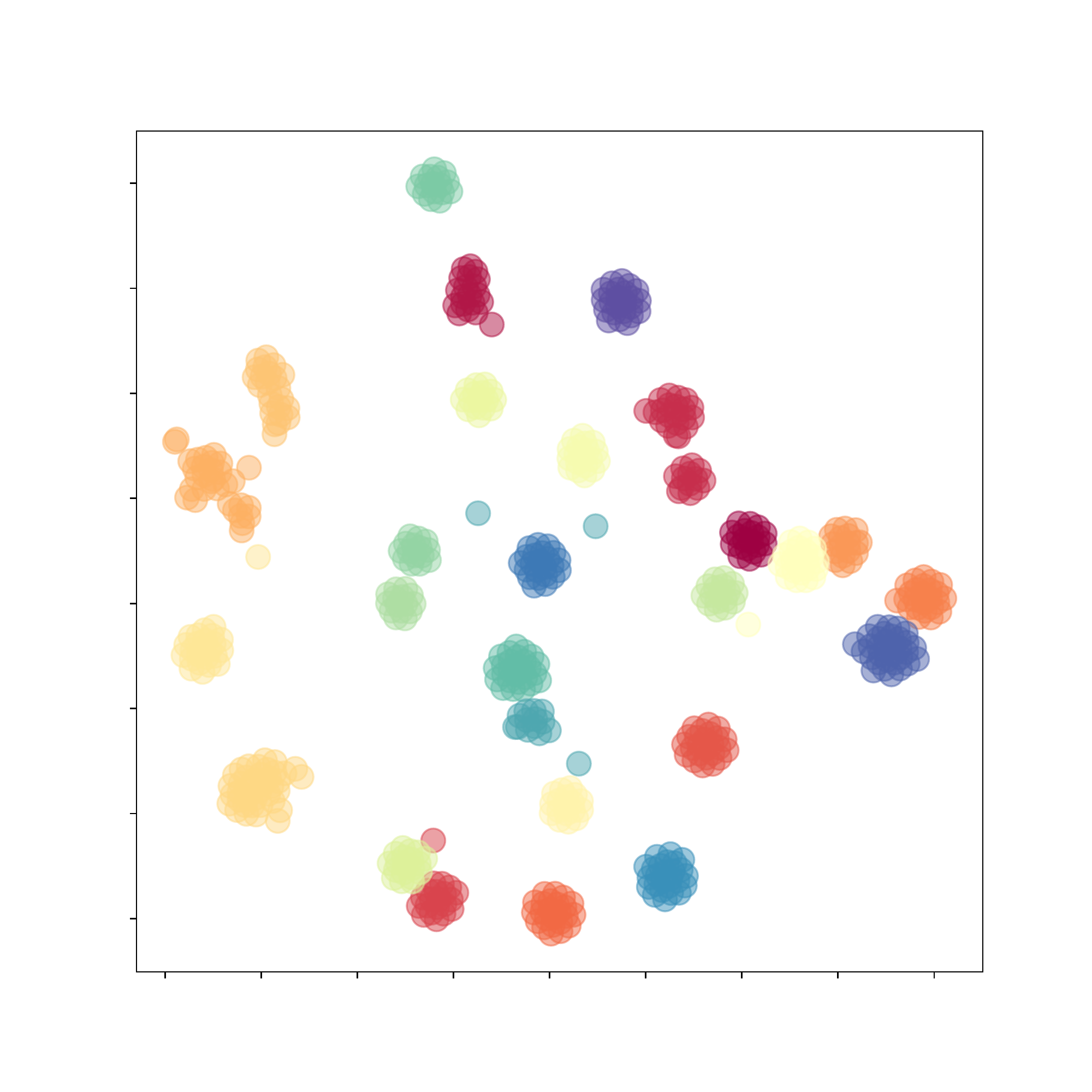}
\label{6-a}
}
\hspace{-7mm}
\subfigure[$z_{sp}^I$ (VSD)]{
\includegraphics[width=0.2725\linewidth]{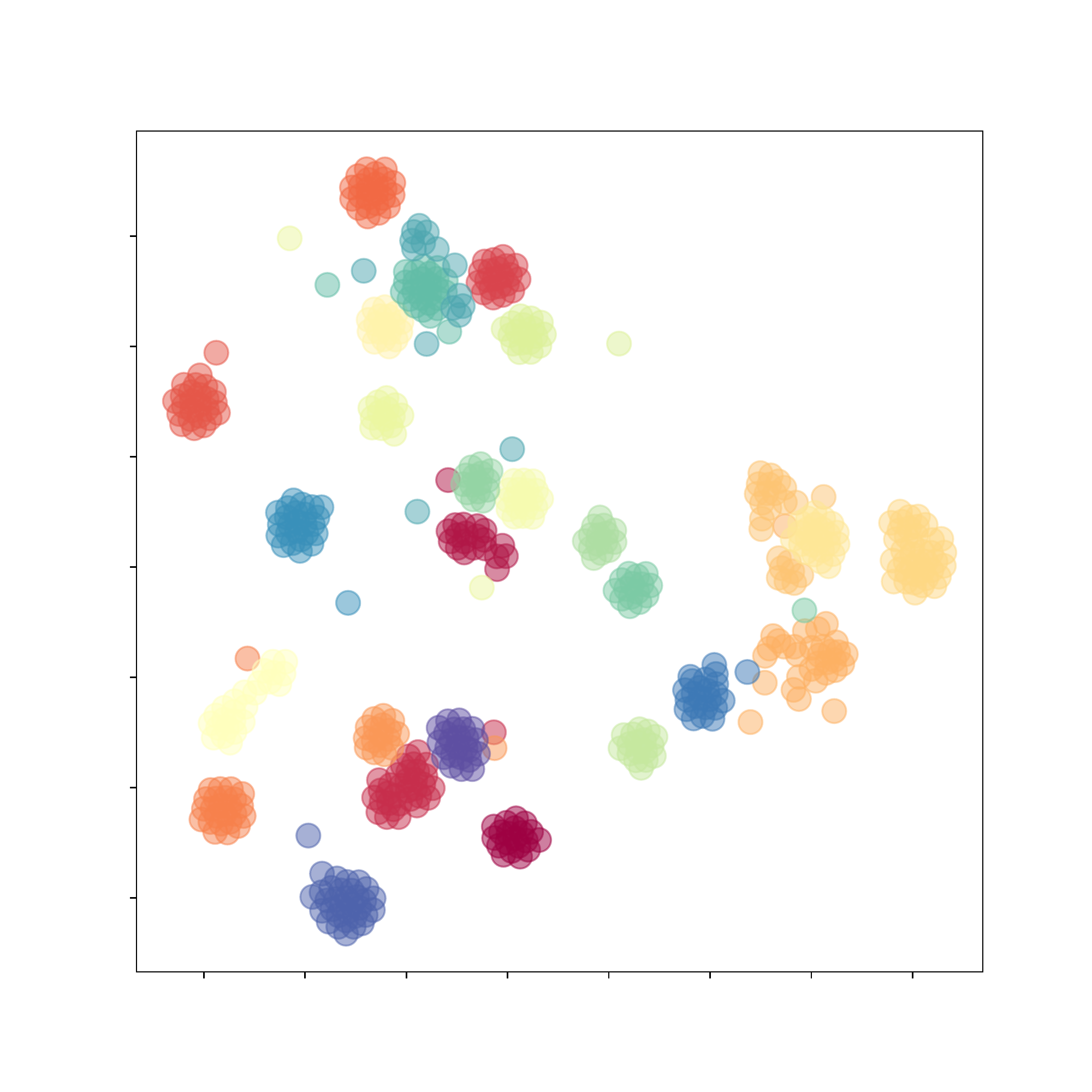}
\label{6-b}	
}
\hspace{-7mm}
\subfigure[$z_{sh}^V$ (VCD)]{
\includegraphics[width=0.2725\linewidth]{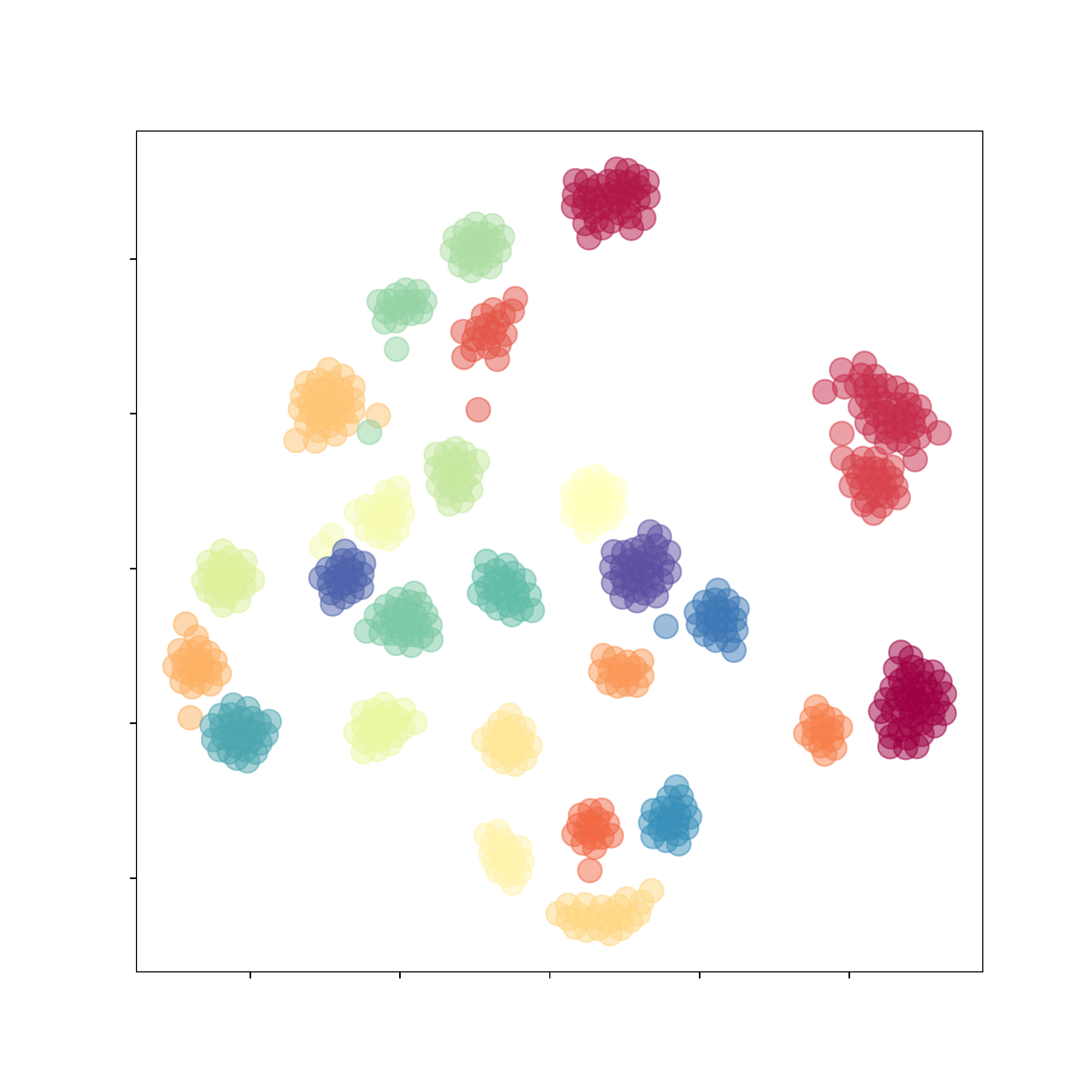}
\label{6-c}
}
\hspace{-7mm}
\subfigure[$z_{sh}^I$ (VCD)]{
\includegraphics[width=0.2725\linewidth]{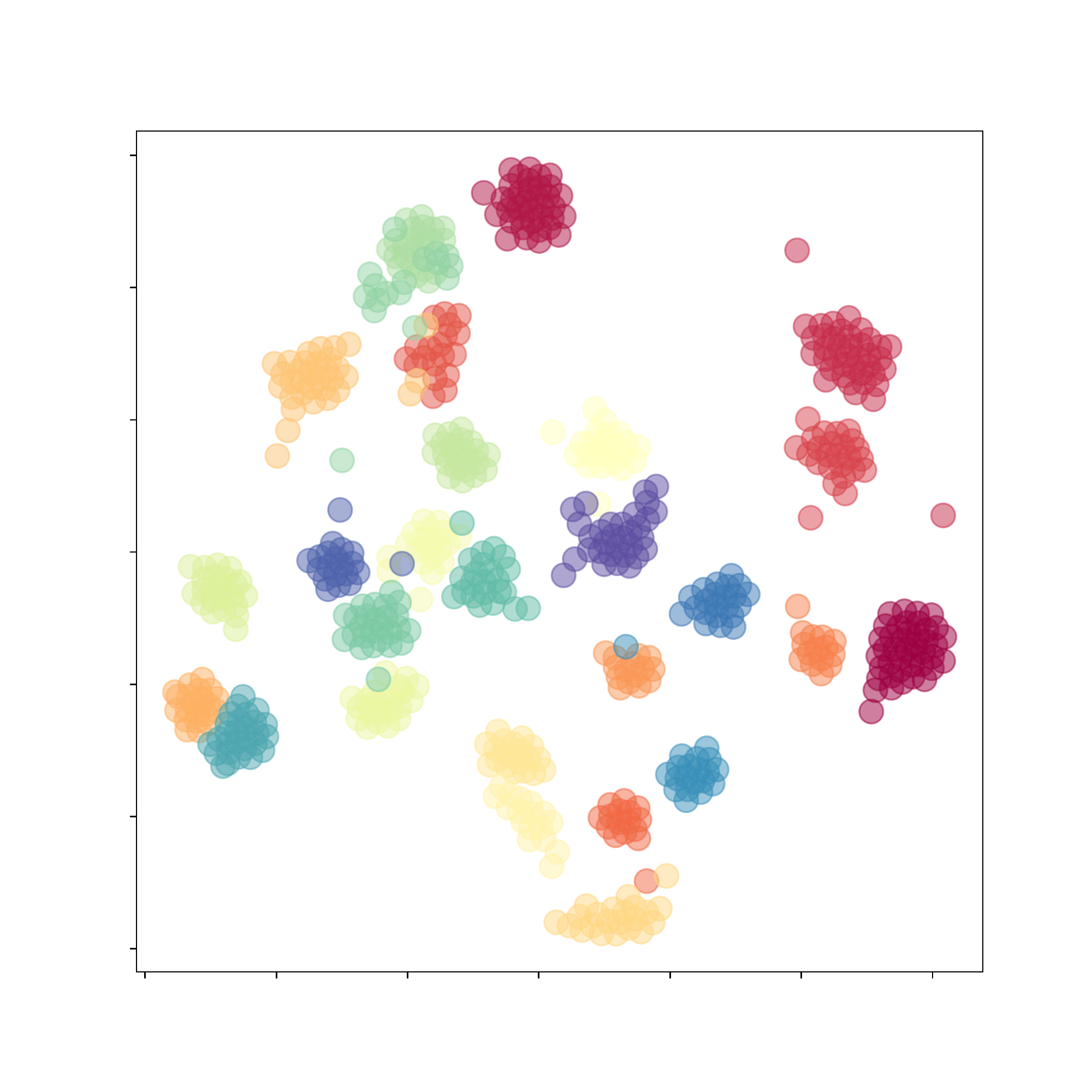}
\label{6-d}	
}
\vspace{-5mm}

\subfigure[$z_{sp}^V$ (CIB)]{
\includegraphics[width=0.2725\linewidth]{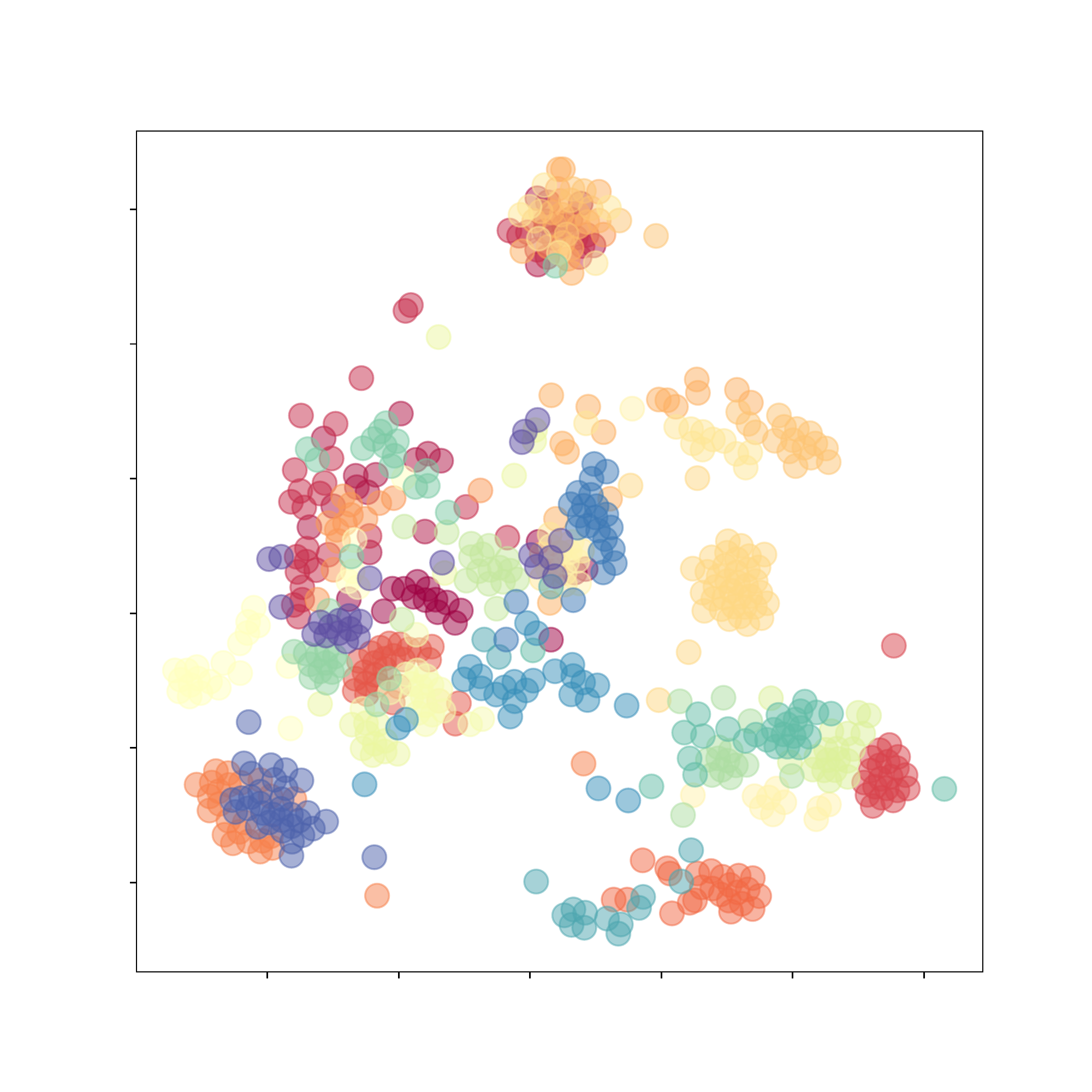}
\label{6-e}	
}
\hspace{-7mm}
\subfigure[$z_{sp}^I$ (CIB)]{
\includegraphics[width=0.2725\linewidth]{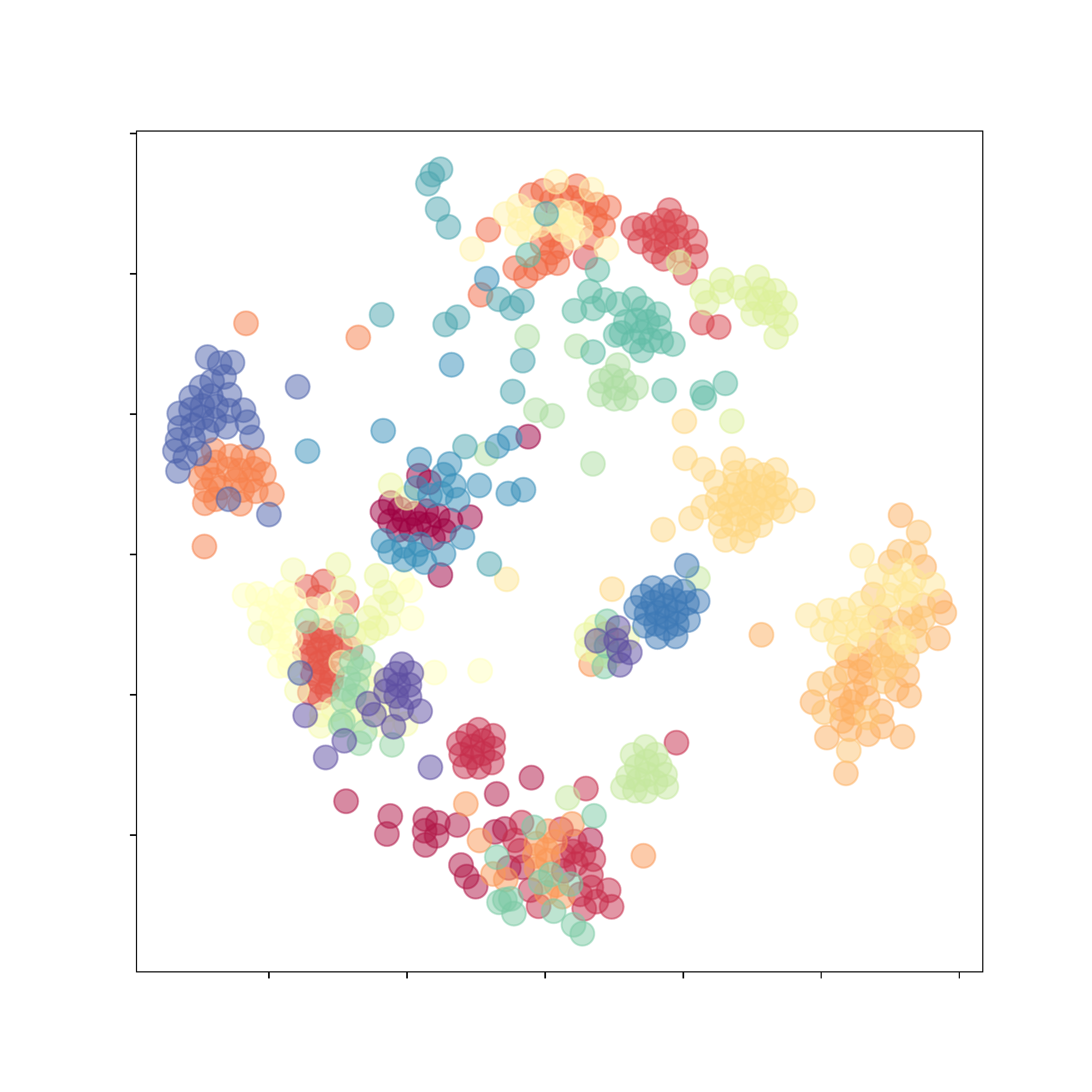}
\label{6-f}
}
\hspace{-7mm}
\subfigure[$z_{sh}^V$ (CIB)]{
\includegraphics[width=0.2725\linewidth]{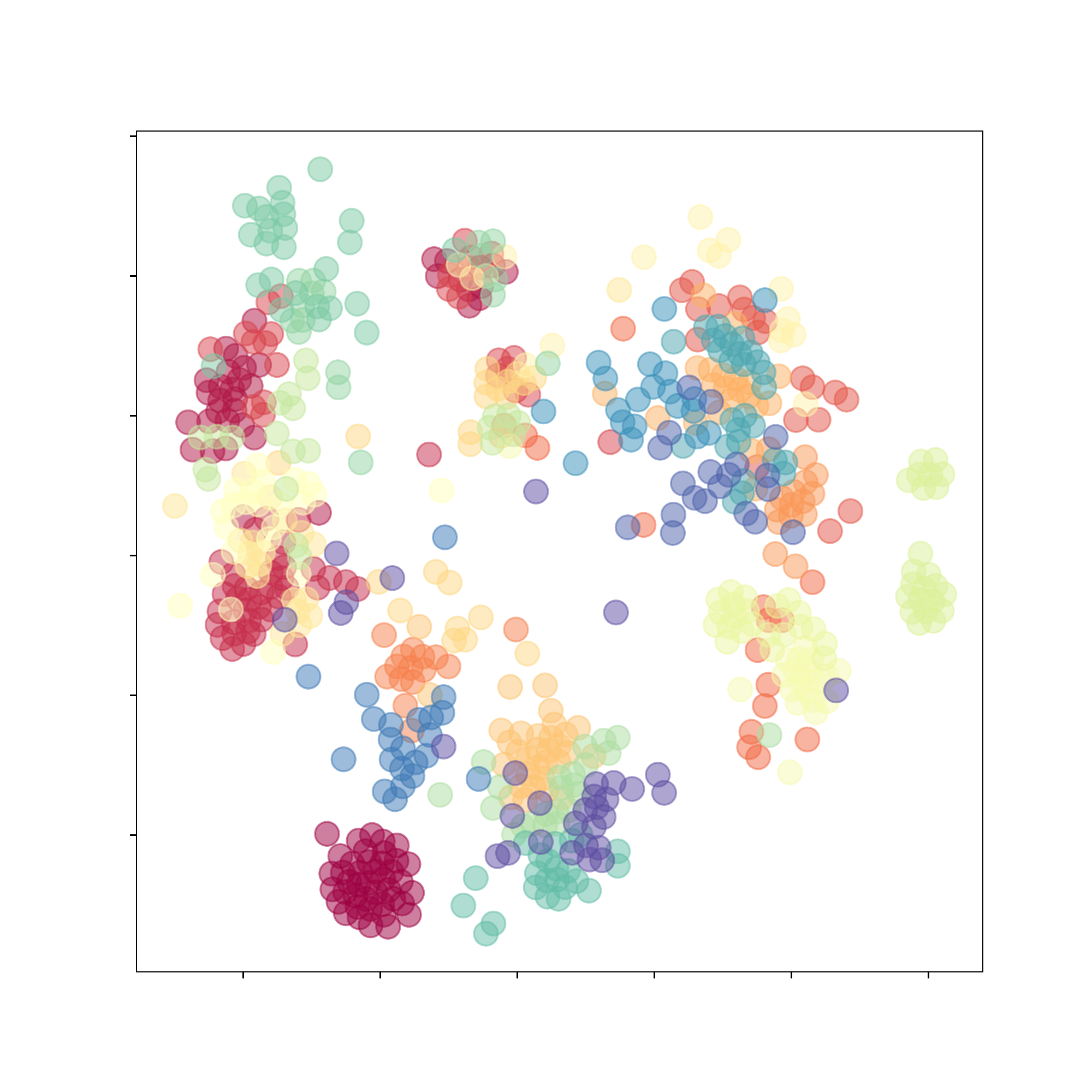}
\label{6-g}	
}
\hspace{-7mm}
\subfigure[$z_{sh}^I$ (CIB)]{
\includegraphics[width=0.2725\linewidth]{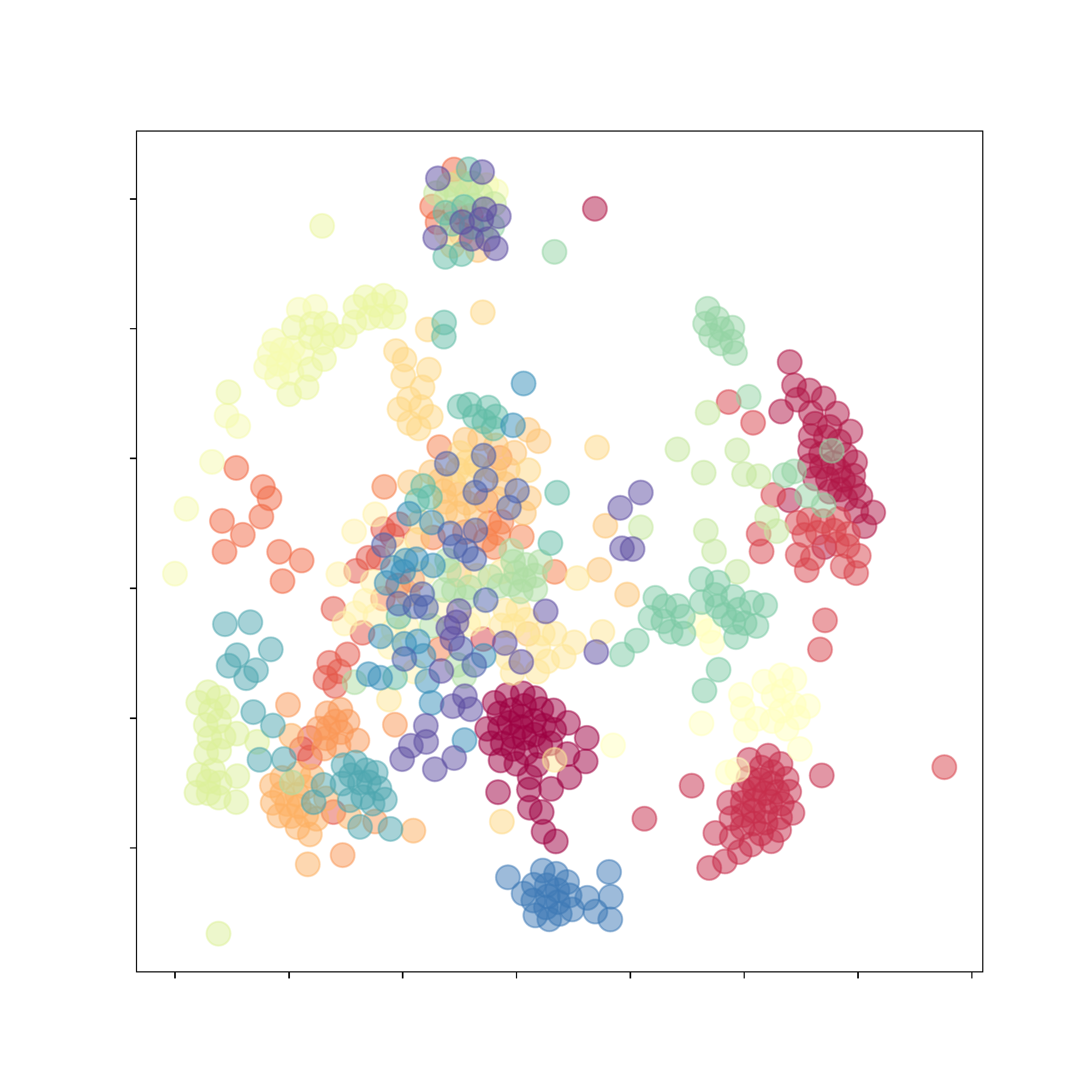}
\label{6-h}	
}
\vspace{-2mm}
\caption{2-D projections of the embedding space by using t-SNE. The results are obtained from our method and conventional IB on SYSU-MM01 test set. Different colors are used to represent different person IDs.}\label{projection}

\end{figure}

\begin{figure}[t]
\centering
\renewcommand{\figurename}{Figure}
\subfigure[$z_{sh}$ (VCD)]{
\includegraphics[width=0.45\linewidth]{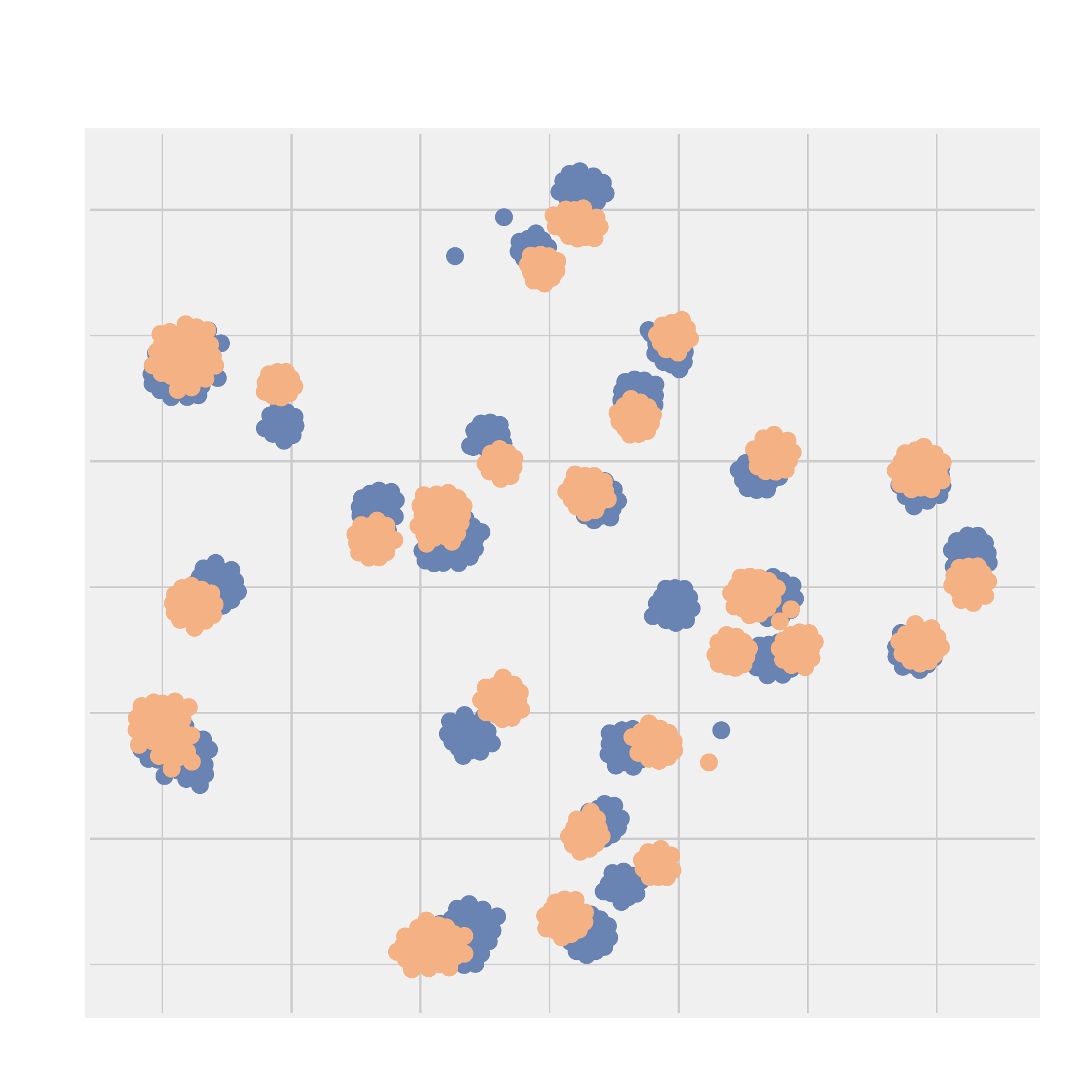}
\label{fusion_vcd}
}
\subfigure[$z_{sh}$ (CIB)]{
\includegraphics[width=0.45\linewidth]{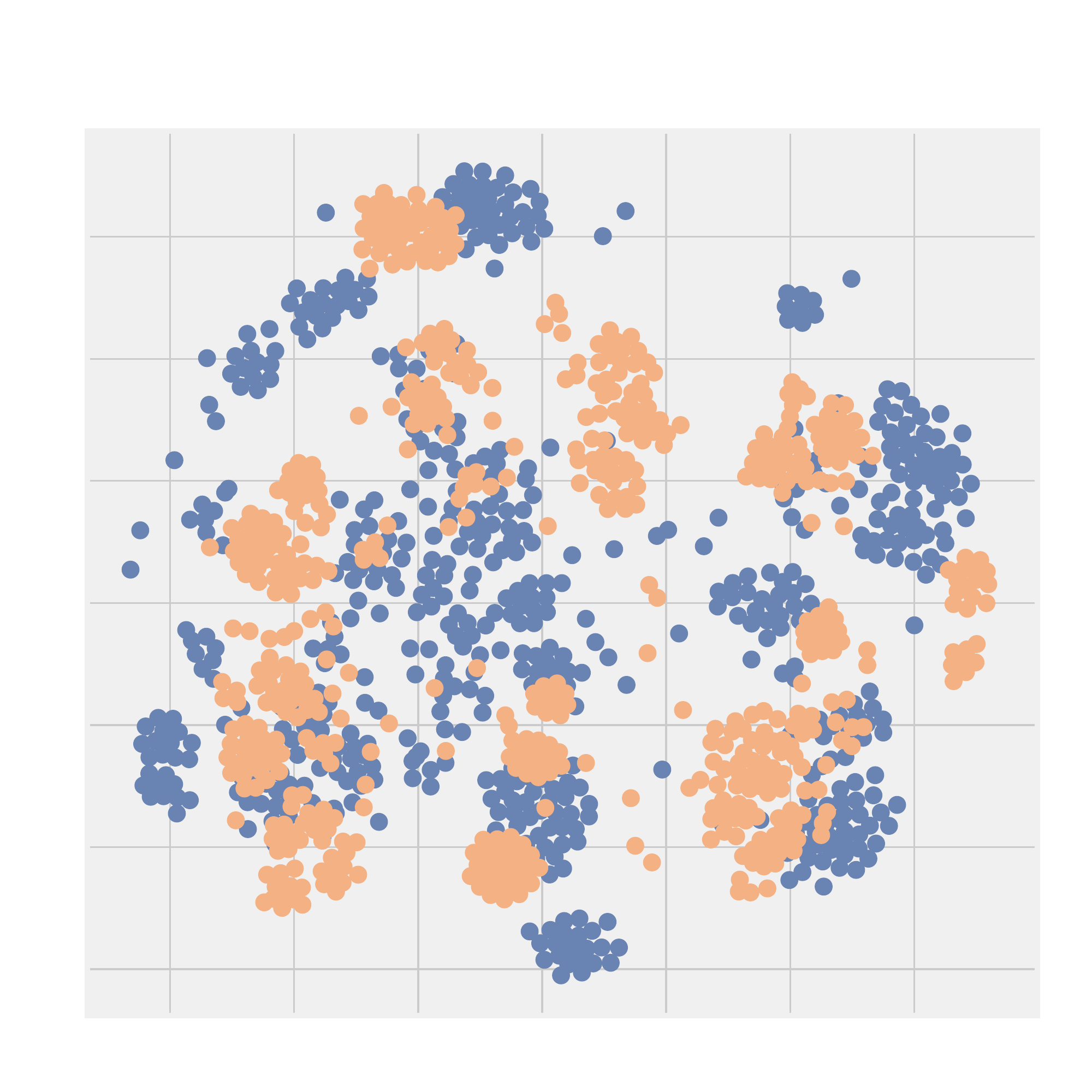}
\label{fusion_cib}
}
\vspace{-2.5mm}
\caption{2D projections of the joint embedding spaces of $z_{sh}^I$ and $z_{sh}^V$ obtained by using t-SNE on SYSU-MM01.}\label{fusion}
\end{figure}

{\textbf{Sufficiency \& Consistency.}} For better illustration, we plot the 2D projection of the representations by using t-SNE on Fig. \ref{projection} and Fig. \ref{fusion}, where we compare our approach with the conventional IB. In particular, $z_{sp}$ and $z_{sh}$ denote the representations obtained from the modal-specific branches and the modal-shared branch respectively, and the superscripts $I$ and $V$ indicate the corresponding inputs are infrared or visible. Based on Fig. \ref{projection} and Fig. \ref{fusion}, we have:

(i) As shown in Fig. \ref{6-e}$\sim$\ref{6-h}, the embedding space of conventional IB is mixed, demonstrating the inferior predictive power and severe redundancy of the learned representation. On the contrary, our method shows evident boost to the discriminative ability with clear class boundaries (see Fig. \ref{6-a}$\sim$\ref{6-d}).

(ii) From Fig. \ref{6-g}, \ref{6-h} and \ref{fusion_cib}, we observe the conventional IB is quite vulnerable and sensitive to modal changes, where we can find the discrepant embedding spaces from different modals. Such phenomenon is not surprising since conventional IB cannot explicitly distinguish modal-consistent/specific information. By contrast, the embedding space of $z^I_{sh}$ and $z^V_{sh}$ obtained from our method appears to coincide with each other (see Fig. \ref{6-c}, \ref{6-d} and \ref{fusion_vcd}), implying that we can learn a consistent representation.

{\textbf{Complexity.}} We also compare the extra computational and memory cost brought by our method and conventional IB. As shown in Tab. \ref{Complexity}, ``\textbf{Enc}'' denotes the encoder, {\it i.e.}, backbone network, ``\textbf{IB}'' and ``\textbf{MIE}'' represents the information bottleneck architecture and mutual information estimator. Clearly, our approach avoids explicit calculation to Eq. (\ref{objective of ib}), and thus implement IB principle with negligible cost.

\subsection{Multi-View Classification}
Multi-view classification aims to optimally integrate various representations from different visual views to improve classification accuracy. ``Multi-view'' in this context means every object is described by different descriptors, and hence there exists a pre-extracted feature set including heterogeneous features with tremendous diversity and complementary information. This scenario therefore is satisfactory to valid the effectiveness of our MV$^2$D (multiple views cases), which allows the network to promote both sufficiency and view-consistency.

\begin{table}
\begin{threeparttable}[t]
\centering
\footnotesize
\caption{Accuracy of our method when using different training strategies. Note all experiments are conducted on SYSU-MM01 under all-search single-shot mode.}\label{effectiveness}
\renewcommand{\arraystretch}{1.25}
\begin{tabular}{c|l|c|c|c|c}
	\hline
	\diagbox & Settings & R1 & R10 &R20 & mAP \\\hline\hline
	1 &$E_{S}$& 53.19 &88.19 &94.69 &49.16 \\
	2{\color{red}\tnote{$\dagger$}} &$E_{S}$+CIB& 38.02 &74.68&83.85 &37.94 \\
	3 &$E_{S}$+$B_{S}$&56.95&93.11 &97.66 &57.01\\
	4 &$E_{S}$+$B_{S}$+VCD+VMD&61.16&94.61&97.97 &60.76 \\\hline\hline
	5 &$E_{I/V}$&57.61&93.68&97.68&56.24\\
	6{\color{red}\tnote{$\dagger$}} &$E_{S/I/V}$+CIB& 41.65 &79.65&88.77&41.69 \\ 
	7 &$E_{I/V}$+$B_{I/V}$&62.26&95.08 &98.79 &59.27 \\
	8 &$E_{I/V}$+$B_{I/V}$+VSD& {\textbf{\color{blue}69.33}} & {\textbf{\color{blue}95.71}}&{\textbf{\color{blue}98.63}} &{\textbf{\color{blue}66.27}} \\\hline\hline
	9 &$E_{S/I/V}$&58.60&93.59 & 97.83 &57.35 \\
	10{\color{red}\tnote{$\dagger$}} &$E_{S/I/V}$+CIB&43.81 &82.91&92.82 &41.77\\
	11 &$E_{S/I/V}$+$B_{S/I/V}$&64.15 &94.42&98.68&61.74 \\
	12 &$E_{S/I/V}$+$B_{S/I/V}$+VD& {\color{red}\textbf{70.02}} & {\color{red}\textbf{96.17}} &{\color{red}\textbf{98.76}}& {\color{red}\textbf{66.70}}\\\hline
\end{tabular}
\vspace{0.9mm}
\vspace{1.6mm}
\begin{tablenotes}
	\footnotesize
	\item[{\color{red}$\dagger$}] Some results are compared for completeness, as conventional IB does not explicitly enforce any constraints to the observation. 
\end{tablenotes}
\vspace{0.9mm}
\vspace{1.6mm}
\vspace{0.9mm}
\vspace{1.6mm}
\end{threeparttable}
\end{table}

\begin{table}[t]
\centering
\renewcommand{\arraystretch}{1.25}
\small
\begin{tabular}{|l|ccc|c|c|}
\hline
Method& Enc & IB& MIE & Time& Params \\ \hline
Baseline& $\surd$ & & & 1.0x& 1.0x\\ 
Ours& $\surd$ & $\surd$ & & 1.09x & 1.15x\\ 
CIB & $\surd$ & $\surd$ & $\surd$ & 1.26x & 1.35x \\ \hline
\end{tabular}
\caption{Computational cost of different methods.}\label{Complexity}
\vspace{-1.6mm}
\vspace{-0.9mm}
\end{table}


\subsubsection{Evaluation Protocol and Benchmarks}
Following \cite{DCCA,DCCAE,MvNNcor}, the adopted multi-view benchmark datasets are split into three parts ({\it i.e.}, $70\%/20\%/10\%$) for training, validation and testing, respectively. Classification accuracy is utilized as the prominent evaluation metric for conducting comparisons with the state-of-the-art techniques.

\begin{figure*}[ht]
\centering
\renewcommand{\figurename}{Figure}
\includegraphics[width=1\linewidth]{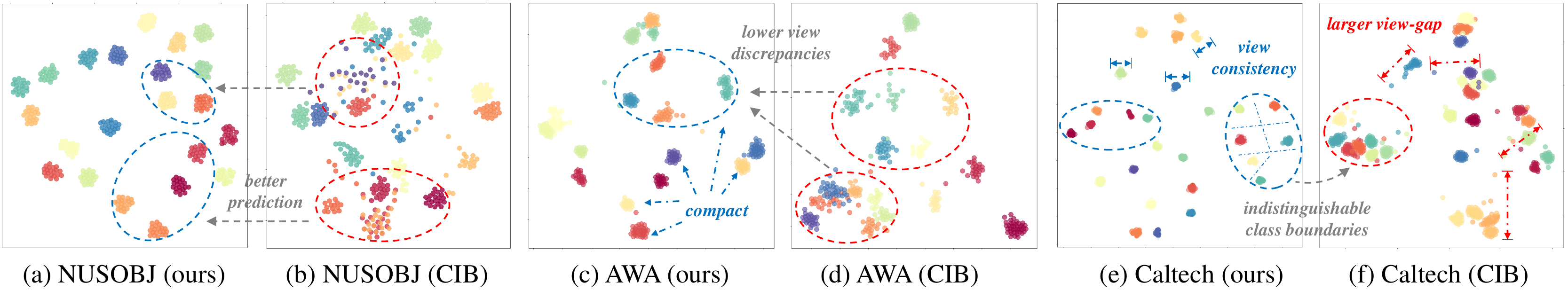}
\vspace{-0.9mm}
\vspace{-1.6mm}
\vspace{-0.9mm}
\vspace{-1.6mm}
\caption{2D projections of the embedding obtained applying MV$^2$D and conventional IB strategy to the NUSOBJ, AWA, Caltech datasets. The representation is projected onto the two principal components, where different colors are used to represent various categories.}
\label{MVD_scatter}
\vspace{-0.9mm}
\vspace{-1.6mm}
\end{figure*}

{\textbf{Caltech-101/20}} \cite{Caltech} consists of $101$ categories of images. Following \cite{Caltech-classification,MvNNcor}, we select the widely used $2386$ images of $20$ classes and $9144$ images of $102$ classes ($101$ object categories and an additional background class), respectively, denoted as Caltech-20 and Caltech-101. This dataset provides 6 kinds of pre-extracted features for each image, {\it i.e.}, $48$-D Gabor, $40$-D Wavelet moments, $254$-D CENTRIST, $1984$-D HOG, $512$-D GIST, and $928$-D LBP.

{\textbf{AWA}} \cite{AWA} is composed of $30,475$ images of $50$ different animals with 6 heterogeneous pre-extracted features for each image. Specifically, they are $2688$-D Color Histogram, $2000$-D Local Self-Similarity, $252$-D Pyramid HOG, $2000$-D SIFT, $2000$-D color SIFT, and $2000$-D SURF.

{\textbf{NUSOBJ}} \cite{NUSOBJ} is a subset of NUS-WIDE and contains $31$ object categories and $30,000$ images in total. It has 5 types of low-dimensional features extracted from all images, including $64$-D color histogram, $225$-D block-wise color moments, $144$-D color correlogram, $73$-D edge direction histogram, and $128$-D wavelet texture.

{\textbf{Reuters}} \cite{Reuters} is a document dataset collected from $5$ different languages. It contains $18,758$ documents in total, all of which are uniformly categorized into $6$ classes. Note different languages can be seen as different views, that is, English ($21,531$-D), French ($24,892$-D), German ($34,251$-D), Italian ($15,506$-D) and Spanish (11547-D).

{\textbf{Hand}} \cite{Hand} consists of features of handwritten numerals extracted from a collection of Dutch utility maps, $200$ patterns per categories (a total of $2000$ patterns). These digits are represented in terms of $6$ feature sets, containing $76$-D Fou, $216$-D Fac, $64$-D Kar, $240$ Pix, $47$-D Zer, and $6$-D Mor.

\begin{table}[t]
\centering
\scriptsize
\renewcommand{\arraystretch}{1.25}
\begin{tabular}{l|c|c|c|c|c|c}
\hline
Method & Cal101 & Cal20& AWA& NUS & Reuters& Hand \\ \hline
SVMcon \cite{SVMcon} & 47.90 & 83.83 & 31.04 & 42.72 & 88.18 & 97.67 \\
DeepLDA \cite{DeepLDA} & 45.65 & 76.51 & 25.60 & 20.32 & 84.91 & 97.67 \\
MvDA \cite{MvDA}& 45.20 & 76.28 & 9.79& 11.46 & 78.83 & 21.33 \\
DCCA \cite{DCCA}& 66.18 & 86.50 & 20.68 & 28.75 & 64.92 & 91.60 \\
DCCAE\cite{DCCAE}& 26.89 & 50.27 & 13.48 & 27.48 & 56.53 & 80.00 \\
GradKCCA \cite{GradKCCA} & 50.53 & 92.92 & 33.33& 48.15 & 43.39 & 95.74\\
MvNNcor\cite{MvNNcor} & 76.00 & 97.92 & 47.69 & 52.05 & 89.28 & {\textbf{\color{blue}99.48}}\\
CPM-Nets \cite{CPM-Nets} & {\textbf{\color{blue}83.22}}& {\textbf{\color{blue}98.23}} & {\textbf{\color{blue}54.38}} & 57.39& {\textbf{\color{blue}93.10}}& {\textbf{\color{red}99.56}}\\\hline
ours (baseline) & 82.66 & 96.87& 54.04 & {\textbf{\color{blue}58.45}} & 93.06 &99.29 \\ 
ours & {\textbf{\color{red}85.93}} & {\textbf{\color{red}99.16}}& {\textbf{\color{red}56.25}} & {\textbf{\color{red}59.60}} & {\textbf{\color{red}95.46}} &99.40 \\ 
\hline
\end{tabular}
\caption{Comparison to the state-of-the-arts on multi-view classification datasets, where the results are obtained using the average of five experiments.}\label{comparison on multi view}
\end{table}

\subsubsection{Implementation Details}

{\textbf{Critical Architectures.}} We choose MvNNcor \cite{MvNNcor} as our baseline, which is composed of two parts, {\it i.e.}, a set of neural networks $\{f_i\}_{v=i}^M$ ({\it i.e.}, the encoder $E_\theta$ in Fig. \ref{MVD_framework}), and an auxiliary module $\{f_{\psi}\}$. Formally, $M$ denotes the total number of viewpoints and each $f_i$ is a fully-connected network consisting of $d_i$ input units and two hidden layers with $512$ and $256$ units equipped with ReLU activation function. To implement MV$^2$D, we append an information bottleneck architecture to each $f_i$ referred to Fig. \ref{MVD_framework}, where we omit $f_{\psi}$ for simplicity (more details can be found in A.3 in the supplementary material).

{\textbf{Training.}} We follow the same experimental configurations in \cite{MvNNcor}, where all experiments are optimized by Adam with $\beta_1= 0.5$ and $\beta_2 = 0.9$. The learning rate is initialized with $10^{-3}$ and decays 20 times at $30$-th and $60$-th epoch. All networks are trained from scratch with a batch size of $64$, and are updated with 160 epochs in total. The training objective includes three terms, {\it i.e.}, classification loss, 
rank loss and Eq. (\ref{refined solution}).

\begin{table}[t]
\centering
\scriptsize
\renewcommand{\arraystretch}{1.25}
\begin{tabular}{l|c|c|c|c|c|c}
\hline
Settings & Cal101 & Cal20& AWA& NUS & Reuters& Hand \\ \hline
$E_{\theta}$ & 81.28 & 94.66& 53.51 & 57.79 & 91.69 &99.20 \\ 
$E_{\theta}$+CIB & 64.29 & 81.16 & 37.90 & 45.88 & 74.39 &87.53 \\ 
$E_{\theta}$+$E_{\phi}$ & {\textbf{\color{blue}82.66}} & {\textbf{\color{blue}96.87}} & {\textbf{\color{blue}54.04}} & {\textbf{\color{blue}58.45}} & {\textbf{\color{blue}93.06}} &{\textbf{\color{blue}99.29}} \\ 
$E_{\theta}$+$E_{\phi}$+MV$^2$D & {\textbf{\color{red}85.93}} & {\textbf{\color{red}99.16}}& {\textbf{\color{red}56.25}} & {\textbf{\color{red}59.60}} & {\textbf{\color{red}95.46}} &{\textbf{\color{red}99.40}} \\ 
\hline
\end{tabular}
\caption{Performance of our approach when adopting different settings.}\label{ablation on multi view}
\end{table}

\begin{figure}[t]
\centering
\renewcommand{\figurename}{Figure}
\vspace{-1.6mm}
\subfigure[Accuracy on AWA with varying dimension of IB.]{
\includegraphics[width=0.45\linewidth]{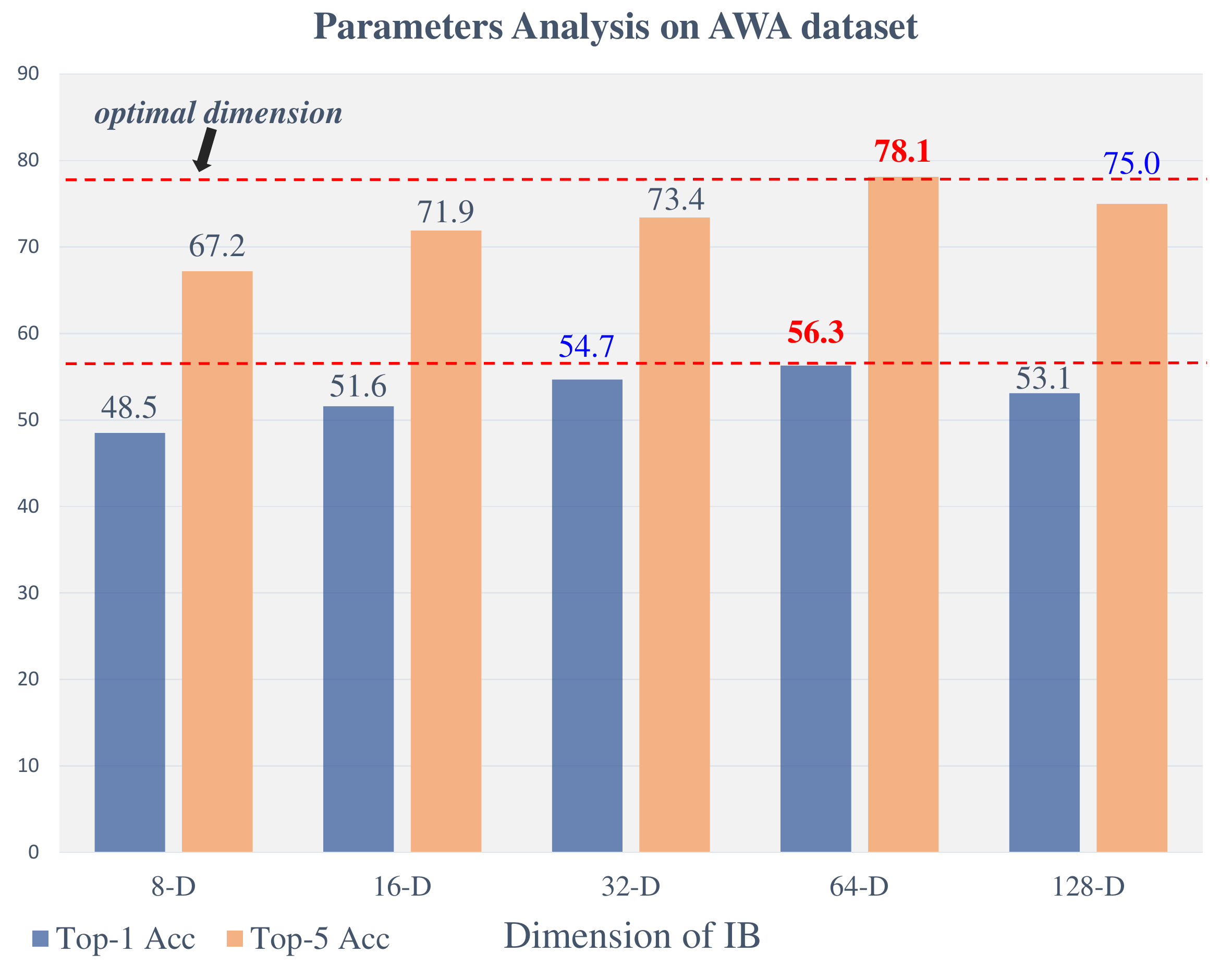}
}
\quad
\subfigure[Accuracy on Caltech with varying dimension of IB.]{
\includegraphics[width=0.45\linewidth]{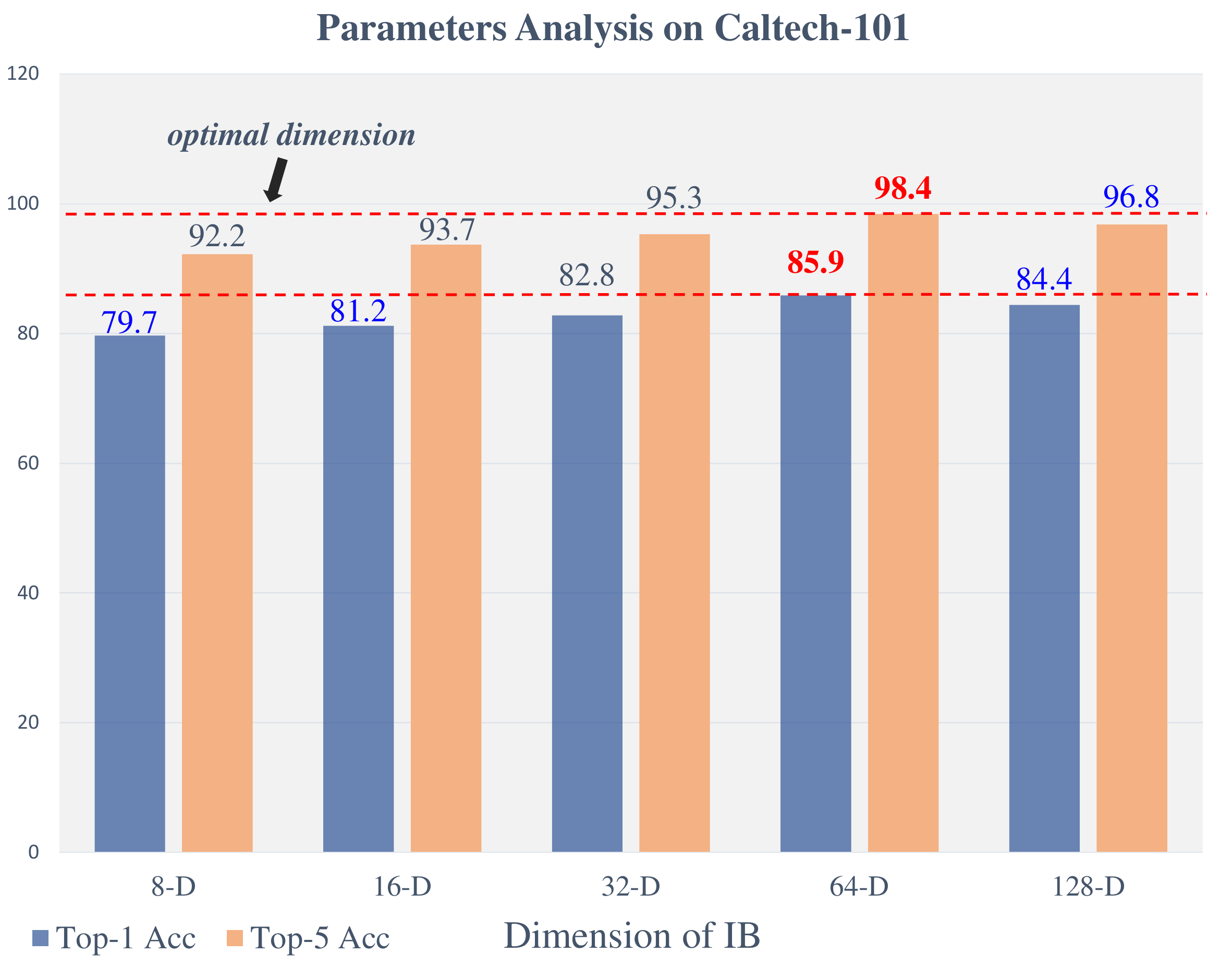}
}
\vspace{-2mm}
\caption{Analysis on the dimension of the information bottleneck. This evaluation is conducted on AWA and Caltech-101 datasets.}\label{analysis on dim}
\vspace{-1.6mm}
\vspace{-0.9mm}
\vspace{-0.9mm}
\end{figure}

\subsubsection{Experimental Results}

\begin{table*}[h]
\centering
\scriptsize
\renewcommand{\arraystretch}{1.5}
\caption{Quantitative results of different approaches on nuScenes\_lidarseg validation set.}	\label{tab:nusc}
\resizebox{\linewidth}{!}{
\begin{tabular}{@{}l|c|cccccccccccccccc}
	\hline
	Methods& \rotatebox{90}{mIoU(\%)}& \rotatebox{90}{barrier} & \rotatebox{90}{bicycle} & \rotatebox{90}{bus} & \rotatebox{90}{car} & \rotatebox{90}{construction} & \rotatebox{90}{motorcycle} & \rotatebox{90}{pedestrian} & \rotatebox{90}{traffic\_cone} & \rotatebox{90}{trailer} & \rotatebox{90}{truck} & \rotatebox{90}{driveable} & \rotatebox{90}{other\_flat} & \rotatebox{90}{sidewalk} & \rotatebox{90}{terrain} & \rotatebox{90}{manmade} & \rotatebox{90}{vegetation} \\ \midrule
	
	RangNet++ \cite{RangeNet++}& 65.5 & 66.0& 21.3& 77.2 & 80.9 & 30.2& 66.8 & 69.6 & 52.1& 54.2& 72.3& 94.1 & 66.6& 63.5 & 70.1& 83.1& 79.8 \\
	SPVCNN \cite{SPVCNN} & 67.8 & 67.1& 12.0& 80.0 & 89.2 & 34.8& 63.5 & 70.0 & 47.0& 48.5& 76.4& 93.6 & 58.6& 67.8 & 72.6& 86.5& 85.4 \\
	PolarNet \cite{PolarNet} & 71.0 & 74.7& 28.2& 85.3 & 90.9 & 35.1& 77.5 & 71.3 & 58.8& 57.4& 76.1& 96.5 & 71.1& 74.7 & 74.0& 87.3& 85.7 \\
	Cylinder3D \cite{Cylinder3D}& 76.1& 76.4& 40.3&91.4 & {\textbf{{\underline{93.8}}}} & 51.3& 78.0 & 78.9 & 64.9& 62.1& {\textbf{{\underline{84.4}}}}& 96.8 & 71.6& {\textbf{{\underline{76.4}}}}& 75.4& 90.5& 87.4 \\
	AF2S3Net \cite{AF2S3Net}&{\color{blue}{\textbf{{\underline{78.3}}}}} & {\textbf{{\underline{78.9}}}} & 52.2 & 89.9 & 84.2 & {\textbf{{\underline{77.4}}}}& 74.3 & 77.3 & 72.0& {\textbf{{\underline{83.9}}}}& 73.8 & {\textbf{{\underline{97.1}}}}& 66.5& 77.5& 74.0& 87.7& 86.8 \\
	PMF \cite{PMF}& 76.9& 74.1& 46.6& 89.8 & 92.1 & 57.0& 77.7 & 80.9 & 70.9& 64.6& 82.9& 95.5 & 73.3& 73.6 & 74.8& 89.4& 87.7 \\	\hline
	\textbf{ours (baseline)} &77.2&74.7& 47.1 &90.0 &92.3&58.1&80.1 &81.4 & 68.6 &62.0 &81.9& 95.6 &73.7 &73.6& {\textbf{{\underline{75.8}}}}& 90.4 & 89.2\\ 
	
	\textbf{ours}&{\color{red}{\textbf{{{\underline{78.9}}}}}} &75.5& \textbf{{\underline{55.8}}} &\textbf{{\underline{93.8}}} &91.7 &61.2&{\textbf{{\underline{83.4}}}} &{\textbf{{\underline{84.0}}}}& \textbf{{\underline{74.2}}} &63.2 &81.3& 95.4 &{\textbf{{\underline{74.2}}}} &73.2& 75.1& \textbf{{\underline{90.6}}} & {\textbf{{\underline{89.4}}}}\\\hline
\end{tabular}

}
\end{table*}
\begin{table*}[h]
\centering
\footnotesize
\renewcommand{\arraystretch}{1.5}
\caption{Quantitative results of different approaches on SemanticKITTI validation set.}\label{tab:semKITTI}
\resizebox{\linewidth}{!}{
\begin{tabular}{@{}l|c|ccccccccccccccccccc@{}}
	\hline
	Methods& \rotatebox{90}{mIoU(\%)}& \rotatebox{90}{road} & \rotatebox{90}{sidewalk} & \rotatebox{90}{parking} & \rotatebox{90}{other ground} & \rotatebox{90}{building} & \rotatebox{90}{car} & \rotatebox{90}{truck} & \rotatebox{90}{bicycle} & \rotatebox{90}{motorcycle} & \rotatebox{90}{other vehicle} & \rotatebox{90}{vegetation} & \rotatebox{90}{trunk} & \rotatebox{90}{terrain} & \rotatebox{90}{person} & \rotatebox{90}{bicyclist} & \rotatebox{90}{motorcyclist} & \rotatebox{90}{fence} & \rotatebox{90}{pole} & \rotatebox{90}{traffic-sign}\\ \midrule
	
	RandLANet \cite{RandLA-Net} & 50.0&90.7 &73.7 &60.2 &20.4 &86.9 &94.2 &40.1 &26.0 &25.8 &38.9 &81.4 &66.8 &49.2 &49.2 &48.2 &7.2 &56.3 &47.7 &38.1 \\
	SPVCNN\cite{SPVCNN}& 58.7& 90.2 &75.4 &67.6 &21.8 &91.6 &97.2 &56.6 &50.6 &50.4 &58.0 &86.1 &{\textbf{{\underline{73.4}}}} &71.0 &67.4 &67.1 &50.3 &66.9 &64.3 &67.3\\
	PolarNet \cite{PolarNet} & 54.3&90.8 &74.4 &61.7 &21.7 &90.0 &93.8 &22.9 &40.3 &30.1 &28.5 &84.0 &65.5 &67.8 &43.2 &40.2 &5.6 &61.3 &51.8 &57.5\\
	BAAF-Net \cite{BAAF-Net} & 59.9&90.9&74.4 &{\textbf{{\underline{62.2}}}} &23.6 &89.8 &95.4 &48.7 &31.8 &35.5&46.7 &82.7 &63.4 &67.9 &49.5 &55.7 &{\textbf{{\underline{53.0}}}} &60.8 &53.7 &52.0\\
	JS3C-Net \cite{JS3C-Net}&{\textbf{\color{blue}{\underline{66.0}}}}&88.9 &72.1 &61.9 &{\textbf{{\underline{31.9}}}} &{\textbf{{\underline{92.5}}}} &95.8 &54.3 &59.3 &{\textbf{{\underline{52.9}}}} &46.0 &84.5 &69.8 &67.9 &69.5 &65.4 &39.9 &{\textbf{{\underline{70.8}}}} &60.7 &68.7\\
	PMF \cite{PMF}&63.9 &{\textbf{{\underline{96.4}}}} &{\textbf{{\underline{80.5}}}} &43.5 &0.1 &88.7 &95.4 &{\textbf{{\underline{68.4}}}} &{\textbf{{\underline{71.6}}}}&0.0 &{\textbf{{\underline{75.2}}}} &{\textbf{{\underline{88.6}}}} &72.7 &{\textbf{{\underline{75.3}}}} &{\textbf{{\underline{78.9}}}} &{\textbf{{\underline{71.6}}}} &0.0 &60.1 &{\textbf{{\underline{65.5}}}} &43.0 \\ \hline
	\textbf{ours (baseline)}&64.7&94.7 &75.8 &55.6 &14.1 &89.4 &96.2 &53.9 &55.7 &50.0&52.5 &86.4 &67.8 &70.4 &63.9 &69.2 &42.5 &63.3 &60.4 &68.1 \\ 
	\textbf{ours} &{\textbf{\color{red}{\underline{66.5}}}}&95.9 &77.4 &60.3 &17.9 &91.6 &{\textbf{{\underline{96.4}}}} &59.1 &56.8 &51.6&53.5 &87.2 &69.7 &71.4 &66.3 &70.5 &46.4 &64.7 &60.2 &{\textbf{{\underline{69.8}}}} \\ \hline
\end{tabular}

}
\end{table*}

{\textbf{Comparison.}} Tab. \ref{comparison on multi view} and Tab. \ref{ablation on multi view} summarize the quantitative results on multi-view classification. Due to the rank metric learning, we obtain a relatively stronger baseline compared with MvNNcor, and MV$^2$D achieves a significant improvement beyond this baseline on all benchmark datasets. Compared with SOTA CCA-based methods \cite{DCCA,DCCAE,GradKCCA}, 
our method also demonstrates promising advantages on the classification performance ({\it e.g.}, outperforms \cite{GradKCCA} by $35.4\%$ on Caltech101). The improvement brought by MV$^2$D can be mostly attributed to the accurate elimination of both non-predictive and view-specific information, which neutralizes sensitivity to view-changes. 
On the other hand, we also observe our information-theoretic constraint drives the deep models \cite{MvNNcor,CPM-Nets,DeepLDA} to learn the sufficient and consistent representations, by achieving stronger performance without requiring complex designs.

{\textbf{Ablation Study.}} Based on the Tab. \ref{ablation on multi view}, 
we can draw the similar conclusions in multi-view case: (i) The appended IB architecture can improve the performance as it introduces additional parameters; (ii) Conventional IB strategy still has no benefits in promoting the accuracy under multi-view cases; (iii) MV$^2$D can evidently boost the performance on all datasets but excluding Hand \cite{Hand}. The reason might be the dimension of feature in this dataset is only 6, which can hardly include rich sources of information. Such phenomenon also reveals the shortcomings of MV$^2$D, {\it i.e.}, incapability to choose the optimal dimension, and becoming mediocre when handling fairly low-dimensional objects.


{\textbf{Analysis on Feature Dimension of IB.}} As is shown in Fig. \ref{analysis on dim}, the accuracy first climbs to a peak with the increase of output dimension of IB, and then degrades. We deduce there are two reasons accounting for this phenomenon: (i) necessary information would be inevitably discarded if the dimension is extremely reduced, which can be concluded from our inferior performance on Hand dataset; (ii) Compact representation are usually beneficial for the downstream tasks.


{\textbf{Sufficiency \& Consistency.}} We also plot the 2D projection of representations by using t-SNE on Fig. \ref{MVD_scatter}, where we compare the representations obtained from our approach and conventional IB. By observing the scatters, we have: (i) The embedding space produced by CIB appear to lack discrimination, where we can spot obvious overlapping within each class and indistinguishable boundaries between different categories; (ii) By contrast, almost all the clusters obtained by MV$^2$D concentrate around a respective centroid, suggesting the sufficiency and view-consistency information are better preserved.

\subsection{LiDAR-RGB Semantic Segmentation}
In this section, we further evaluate the variational distillation framework on LiDAR-RGB semantic segmentation, which, in practice, is a typical cross-modal learning problem. It is a fundamental task for scene perception and understanding, which aims to predict a dense label map by fusing complementary information from both LiDAR and RGB sensors. Thus, it is also quite suitable to evaluate the variational distillation framework in such a scalable and complex representation learning problem.



\begin{figure*}[ht]
\centering
\renewcommand{\figurename}{Figure}

\includegraphics[width=1\linewidth]{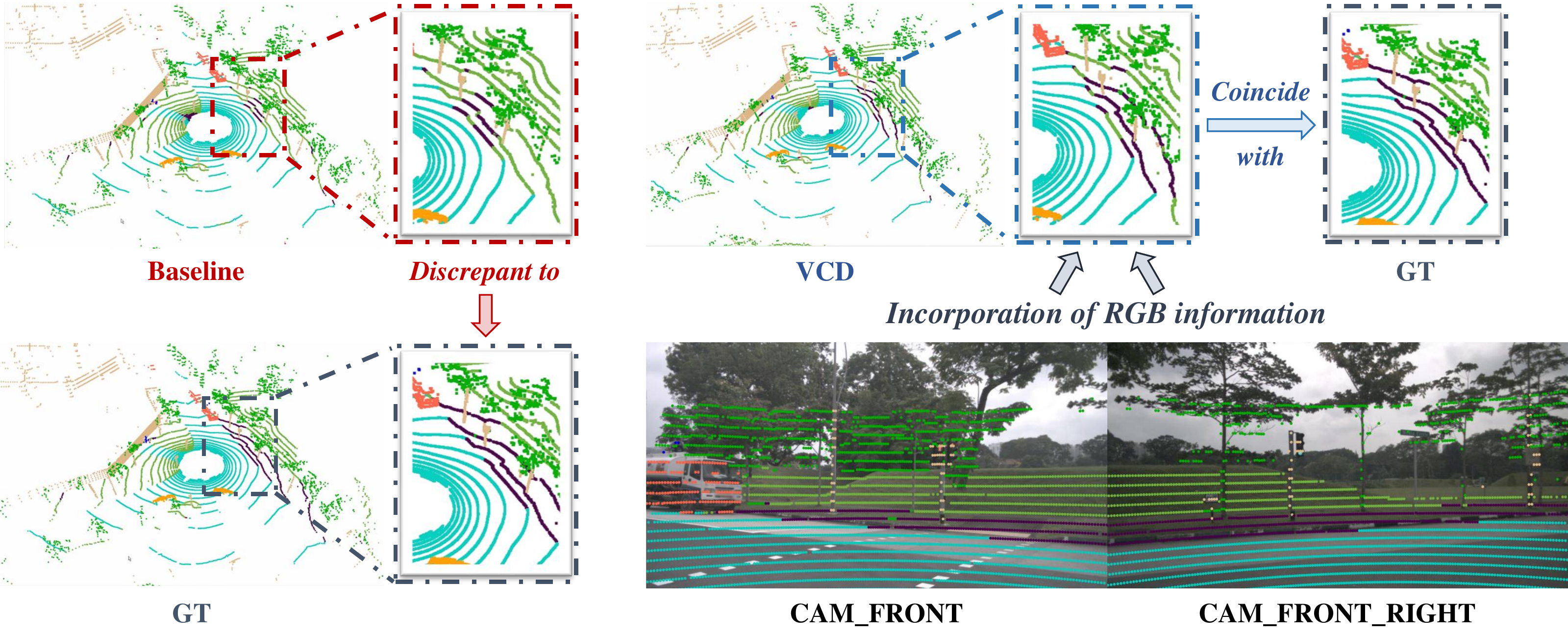}\caption{Qualitative results obtained from our baseline and VCD on nuScenes dataset, where different colors are utilized to denote various categories. Note the RGB images are taken from the front camera and the front-right camera, and provide crucial complementary information for the LiDAR-based segmentation.}
\label{seg: visualization}
\end{figure*}

\subsubsection{Evaluation Protocol and Benchmarks}
To evaluate the proposed method, we follow the official protocol \cite{nuScenes,SemanticKITTI} to leverage mean intersection-over-union (mIoU) as the evaluation metric. For a given class $i$, IoU is formulated as: $I o U_{i}=T P_{i}/(T P_{i}+F P_{i}+F N_{i})$, where ${TP}_i$, ${FP}_i$, ${FN}_i$ represent true positive, false positive, and false negative predictions for the $i$-th class and the mIoU is the mean value of IoU over all classes.

{\textbf{nuScenes}} \cite{nuScenes} collects $1000$ scenes of $20$s duration with $32$ beams LiDAR sensor. The number of total frames is $40,000$, and are split into $28,130$ training frames and $6,019$ validation frames. After merging similar classes and removing rare classes, total 16 classes for the LiDAR semantic segmentation are remained. Unlike SemanticKITTI, which provides only the images of the front-view camera, nuScenes has 6 cameras for different views of LiDAR.

{\textbf{SemanticKITTI}} \cite{SemanticKITTI} is a large-scale driving scene dataset for point cloud segmentation, which provides $43,000$ scans with point-wise semantic annotation. This dataset consists of 22 sequences in total, splitting sequences $00$ to $10$ as training set (where sequence $08$ is used as the validation set), and sequences $11$ to $21$ as test set. $19$ classes are used for training and evaluation after ignoring and merging the classes with very few points or with different moving status.

\subsubsection{Implementation Details}
{\textbf{Critical Architectures.}} Our framework is mainly composed of two sub-networks to handle inputs from different modals ({\it i.e.}, image and LiDAR), and each of which includes a backbone network ({\it i.e.}, $E_{\theta}$) and an information bottleneck architecture to implement VCD. In addition, we also adopt a LI-fusion module \cite{EPNet} to enhance the LiDAR point representation by incorporating image features at multiple scales. More specifically, the image sub-network extracts the semantic information with a set of convolutional operations, which are implemented by SwiftNet \cite{SwiftNet} pretrained on ImageNet. We adopt SPVCNN \cite{SPVCNN} as the Point Cloud sub-network, which outputs the final representation for segmentation. Details and graphical illustration of our framework can be found in A.2 in the supplementary material.


{\textbf{Training.}} All experiments are optimized by SGD with Nesterov, where weight decay and momentum are set to $1\times 10^{-4}$ and $0.9$, respectively. The learning rate starts at $2.4\times10^{-1}$ and adopts the warm-up with cosine scheduler. We train our model for $40$ epochs in total with batch size fixed to $8$, and we conduct all experiments on NVIDIA RTX A6000 GPUs. In addition to the widely adopted cross-entropy, the training objective also consists of multi-class focal loss \cite{focal_loss}, Lov$\acute{{\text{a}}}$sz-softmax loss \cite{Lovasz_loss} and our VCD.

\subsubsection{Experimental Results}
{\textbf{Quantitative Analysis.}} Tab. \ref{tab:nusc} and Tab. \ref{tab:semKITTI} shows the comparison on the validation set of nuScenes and SemanticKITTI. We can draw the following conclusion: Our approach evidently boosts the performance and outperforms the baseline and other competitors in term of mIoU on both benchmark datasets. More specifically, the proposed variational distillation framework exceeds the SPVCNN \cite{SPVCNN} (our point cloud baseline) by a large margin, and it also demonstrates superiority to the projection-based \cite{RandLA-Net,PolarNet,RangeNet++}, voxel partition and 3D convolutional methods \cite{Cylinder3D,BAAF-Net,JS3C-Net,AF2S3Net} on both datasets, revealing its effectiveness. Besides, since \cite{AF2S3Net} and \cite{JS3C-Net} adopt stronger baselines and various modules specifically designed for point cloud, the performance disparities are relatively inconspicuous compared with our method. 



{\textbf{Qualitative Analysis.}} Fig. \ref{seg: visualization} provides a visual illustration to the produced labeling map. Obviously, VCD can better facilitate the fusion of complementary information and thus attains preferable segmentation result. By comparison, we observe some categories which are hardly recognized are ignored by the baseline, which shows IB can handle the huge modal-discrepancy.


\section{Conclusion}
In this work, we provide an analytical solution to fitting mutual information by using variational inference, rather than designing a sophisticated estimator. On this basis, we reformulate the objective of IB, and propose a generalized variational distillation framework, which enables us to jointly preserve the sufficiency of representations and get rid of task-irrelevant distractors. Its special cases, {\it i.e.,} Multi-View Variational Distillation (MV$^2$D), Variational Cross-Distillation (VCD) and Variational Mutual-Distillation (VMD), can produce view-consistent representations among multiple heterogeneous data observations. The future works would include learning an adaptive method to determine the output dimension of IB. Also, more broader multi-view applications such as medical and text would be studied.

\appendices
\section{Proof Details}\label{appendix_c}
	
	Given $\{v_1,v_2,...,v_n\}$ as $n$ observations of the same object $x$ from different viewpoints, domains, or modals, and let $y$ be the ground-truth label. Consider $\{z_1,z_2,...,z_n\}$ to be the corresponding representations obtained from an information bottleneck, we make the following two simple assumptions:
	\\\\
	{\textit{Hypothesis}}:
	\\\\
	({\textit{$H_1$}}) information shared by more views leads to better robustness
	\\\\
	({\textit{$H_2$}}) no representations are only composed of view-specific information
	\\\\
	{\textit{Thesis}}:
	\\\\
	({\textit{$T_1$}}) minimizing $D_{KL}[\mathbb{P}_{z_{\{1,...n\}}} || \mathbb{P}_{z_{\{1,...n\}/i}}]$ is consistent with the objective of eliminating view-specific information, which also complies with sufficiency constraint
	\\\\
	({\textit{$T_2$}}) MV$^2$D automatically and accurately prioritizes different compositions of the preserved information based on the generalization ability
	\subsection{Proof to Thesis 1}\label{proof_1}
	
	Consider $z_i$ as the representation of $v_i$, we have the following factorization using the chain rule \cite{mib,sufficiency}:
	\begin{equation}
		I(v_i;z_i)= \underbrace{I(y;z_i)}_{\operatorname{predictive}}+\underbrace{I(v_i;z_i|y)}_{\operatorname{superfluous}}. \label{MVD_1}
	\end{equation}
	Notice $I(y;z_i)$ is composed of various terms when multiple views are involved (see Fig. \ref{Venn_MVD} for visualization). Hence, we have:
	\begin{equation}
		I(y;z_i)=\underbrace{I_i^c}_{\operatorname{consistent}}+\underbrace{I(y;z_i|z_{\{1,...,n\}/i})}_{\operatorname{view-specific}},\label{MVD_2}
	\end{equation}
	where $I_i^c$ is utilized to uniformly represent all compositions of the view-consistent information encoded in $z_i$, and $I(y;z_i|z_{\{1,...,n\}/i})$ denotes the information that is unique to $z_i$, and is inaccessible to all other representations, {\it i.e.}, view-specific information. Substituting Eq. (\ref{MVD_2}) into Eq. (\ref{MVD_1}), we have:
	\begin{equation}
		I(v_i;z_i)=\underbrace{I(v_i;z_i|y)}_{\operatorname{superfluous}}+\underbrace{I(y;z_i|z_{\{1,...,n\}/i})}_{\operatorname{view-specific}}+\underbrace{I^c_i}_{\operatorname{consistent}}. \label{MVD_2_2}
	\end{equation}
	According to Definition 2, consistency requires to eliminate both $I(v_i;z_i|y)$ and $I(y;z_i|z_{\{1,...,n\}/i})$ while simultaneously maximizing $I^c_i$. However, $I^c_i$ includes multiple terms and is almost possible to be directly optimized. To resolve this issue, we first introduce the following inequality based on the information processing principle:
	\begin{equation}
		I(v_i;z_i)\le I(v_i;z_i|y)+I(y;z_i|z_{\{1,...,n\}/i})+I(y;z_i),\label{MVD_3}
	\end{equation}
	which indicates promoting consistency undergoes several sub-processes: (i) maximizing $I(y;z_i)$; (ii) approximating $I^c_i$ to its upper bound, {\it i.e.}, $I(y;z_i)$; (iii) discarding the task-irrelevant nuisances $I(v_i;z_i|y)$; (iv) eliminating view-specific information $I(y;z_i|z_{\{1,...,n\}/i})$.
	
	To that end, the training objective can be formulated as:
	\begin{equation}
		\min \sum_{i \in n} \underbrace{I(v_i;z_i|y)-I(y;}_{\operatorname{sufficiency}}\overbrace{z_i)+I(y ; z_i | z_{\{1,... ,n\}/i})}^{ \operatorname{consistency}},\label{MVD_4}
	\end{equation}
	where, as elaborated in Sec. \ref{MVD} and Sec. \ref{example1}, the sufficiency term in Eq. (\ref{MVD_4}) accounts for (i) and (iii), while the consistency term are utilized for (ii) and (iv). 
	
	Next, given $z_i\in\{z_1,...,z_n\}$, we have the view-specific information divided as follows \cite{sufficiency,conditional_MI}:
	\begin{flalign} 
		&I\left(y ; z_{i} | z_{\{1,..., n\} / i}\right)=H(y | z_{\{1,..., n\}/i})-H(y | z_{\{1,..., n\}})=& \nonumber\\
		&-\int p(y | z_{\{1,..., n\}/i})\log p(y | z_{\{1,..., n\}/i})~dy\nonumber\\
		&+\int p(y | z_{\{1,..., n\}})\log p(y | z_{\{1,..., n\}})~dy=\nonumber\\
		&-\int p(y | z_{\{1,..., n\}/i})\log \left[\frac{p(y | z_{\{1,..., n\}/i})}{p(y | z_{\{1,..., n\}})}p(y | z_{\{1,..., n\}})\right]dy\nonumber\\
		&+\int p(y | z_{\{1,..., n\}})\log \left[\frac{p(y | z_{\{1,..., n\}})}{p(y | z_{\{1,..., n\}/i})}p(y | z_{\{1,..., n\}/i})\right]dy,\label{MVD_5}
	\end{flalign}
	in which $H(\cdot)$ represents Shannon entropy. By factorizing the first integral in Eq. (\ref{MVD_5}), we obtain:
	\begin{flalign} 
		&\int p(y | z_{\{1,..., n\}/i})\log \left[\frac{p(y | z_{\{1,..., n\}/i})}{p(y | z_{\{1,..., n\}})}p(y | z_{\{1,..., n\}})\right]dy\nonumber\\
		&=\underbrace{\int p(y | z_{\{1,..., n\}/i})\log \left[\frac{p(y | z_{\{1,..., n\}/i})}{p(y | z_{\{1,..., n\}})}\right]dy}_{\operatorname{term~\mathbb{Z}_1}}\nonumber\\
		&+\underbrace{\int p(y | z_{\{1,..., n\}/i})\log p(y | z_{\{1,..., n\}})~dy}_{\operatorname{term ~\mathbb{Z}_2}}.
	\end{flalign}
	Similarly, we have the second one divided as:
	\begin{flalign} 
		&\int p(y | z_{\{1,..., n\}})\log \left[\frac{p(y | z_{\{1,..., n\}})}{p(y | z_{\{1,..., n\}/i})}p(y | z_{\{1,..., n\}/i})\right]dy\nonumber\\
		&=\underbrace{\int p(y | z_{\{1,..., n\}})\log \left[\frac{p(y | z_{\{1,..., n\}})}{p(y | z_{\{1,..., n\}/i})}\right]dy}_{\operatorname{term~\widehat{\mathbb{Z}}_1}}\nonumber\\
		&+\underbrace{\int p(y | z_{\{1,..., n\}})\log p(y | z_{\{1,..., n\}/i})~dy}_{\operatorname{term~\widehat{\mathbb{Z}}_2}}.
	\end{flalign}
	Integrating term $\mathbb{Z}_1$ and term $\widehat{\mathbb{Z}}_1$ over $y$:
	\begin{flalign}
		\mathbb{Z}_1=D_{KL}\left[ p(y | z_{\{1,..., n\}/i}) || p(y | z_{\{1,..., n\}}) \right],\\
		\widehat{\mathbb{Z}}_1=D_{KL}\left[ p(y | z_{\{1,..., n\}}) || p(y | z_{\{1,..., n\}/i}) \right],
	\end{flalign}
	where $D_{KL}[\cdot||\cdot]$ denotes the relative entropy, {\it i.e.}, KL-divergence. Then we integrate term $\mathbb{Z}_2$ and term $\widehat{\mathbb{Z}}_2$ over $y$ and show the following:
	\begin{flalign}
		\mathbb{Z}_2=-H\left( p(y | z_{\{1,..., n\}/i}) , p(y | z_{\{1,..., n\}}) \right),\\
		\widehat{\mathbb{Z}}_2=-H\left( p(y | z_{\{1,..., n\}}) , p(y | z_{\{1,..., n\}/i}) \right).\label{MVD_6}
	\end{flalign}
	Obviously, both $\mathbb{Z}_2$ and $\widehat{\mathbb{Z}}_2$ are cross entropies.
	
	Based on the above analysis, the view-specific information contained in $I(y;z_i | z_{\{1,...,n\}/i})$ can be represented with:
	\begin{flalign}
		I(y;z_i | z_{\{1,...,n\}/i})= -(\mathbb{Z}_1+\mathbb{Z}_2)+(\widehat{\mathbb{Z}}_1+\widehat{\mathbb{Z}}_2).\label{MVD_7}
	\end{flalign}
	Using the non-negativity of entropies, we have
	\begin{flalign}
		I(y;z_i | z_{\{1,...,n\}/i})\le& D_{KL}\left[ p(y | z_{\{1,..., n\}}) || p(y | z_{\{1,..., n\}/i}) \right] \nonumber\\
		+&H\left( p(y | z_{\{1,..., n\}/i}) , p(y | z_{\{1,..., n\}}) \right).
	\end{flalign}
	Denoting $p(y | z_{\{1,..., n\}/i})$ and $p(y | z_{\{1,..., n\}})$ as $\mathbb{P}_{Z}$ and $\widehat{\mathbb{P}}_{Z}$ for simplicity, we have the upper bound as:
	\begin{flalign}
		\mathbb{E}_{v_i\sim E_{\theta}(v_i|x)} \mathbb{E}_{z_i\sim E_{\phi}(z_i|x)} \left[D_{KL}[ \widehat{\mathbb{P}}_{Z}||\mathbb{P}_{Z}]+H(\mathbb{P}_{Z},\widehat{\mathbb{P}}_{Z})\right],
	\end{flalign}
	where $\theta$ and $\phi$ parameterize the encoder and the information bottleneck. In the view of above, the objective of eliminating view-specific information can be formalized as:
	\begin{flalign}
		\min_{\theta, \phi}\mathbb{E}_{v_i\sim E_{\theta}(v_i|x)} \mathbb{E}_{z_i\sim E_{\phi}(z_i|x)} \left[D_{KL}[ \widehat{\mathbb{P}}_{Z}||\mathbb{P}_{Z}] + H(\mathbb{P}_{Z},\widehat{\mathbb{P}}_{Z})\right].\label{MVD_8}
	\end{flalign}
	Clearly, the objective of eliminating view-specific information is consistent with reducing the discrepancy between $\mathbb{P}_{Z}$ and $\widehat{\mathbb{P}}_{Z}$. Notice this can be attained by minimizing $D_{KL}[ \widehat{\mathbb{P}}_{Z}||\mathbb{P}_{Z}]$ only, which approximates $\mathbb{P}_{Z}$ to $\widehat{\mathbb{P}}_{Z}$ and forces $E_{\phi}$ to focus more on the view-consistent information, preventing explicit violation to the sufficiency constraint. Ideally, $D_{KL}[ \widehat{\mathbb{P}}_{Z}||\mathbb{P}_{Z}]$ is reduced to zero, which results in the coincidence between $\mathbb{P}_{Z}$ and $\widehat{\mathbb{P}}_{Z}$, indicating all predictive cues with view-consistency are preserved by the representation, and we have:
	\begin{flalign}
		\lim _{D_{KL}[ \widehat{\mathbb{P}}_{Z}||\mathbb{P}_{Z}]\rightarrow0} I(y;z_i | z_{\{1,...,n\}/i}) = 0
	\end{flalign}
	Based on Eq. (\ref{MVD_2}), we show:
	\begin{flalign}
		\lim _{D_{KL}[ \widehat{\mathbb{P}}_{Z}||\mathbb{P}_{Z}]\rightarrow0} I(y;z_i)-I^c_i= 0,
	\end{flalign}
	demonstrating that minimizing $D_{KL}[\mathbb{P}_{z_{\{1,...n\}}} || \mathbb{P}_{z_{\{1,...n\}/i}}]$ is consistent with the objective of eliminating view-specific information. Based on the above analysis, ({\textit{$T_1$}}) is proved.
	
	\begin{figure}[t]
		\centering
		\includegraphics[width=1\linewidth]{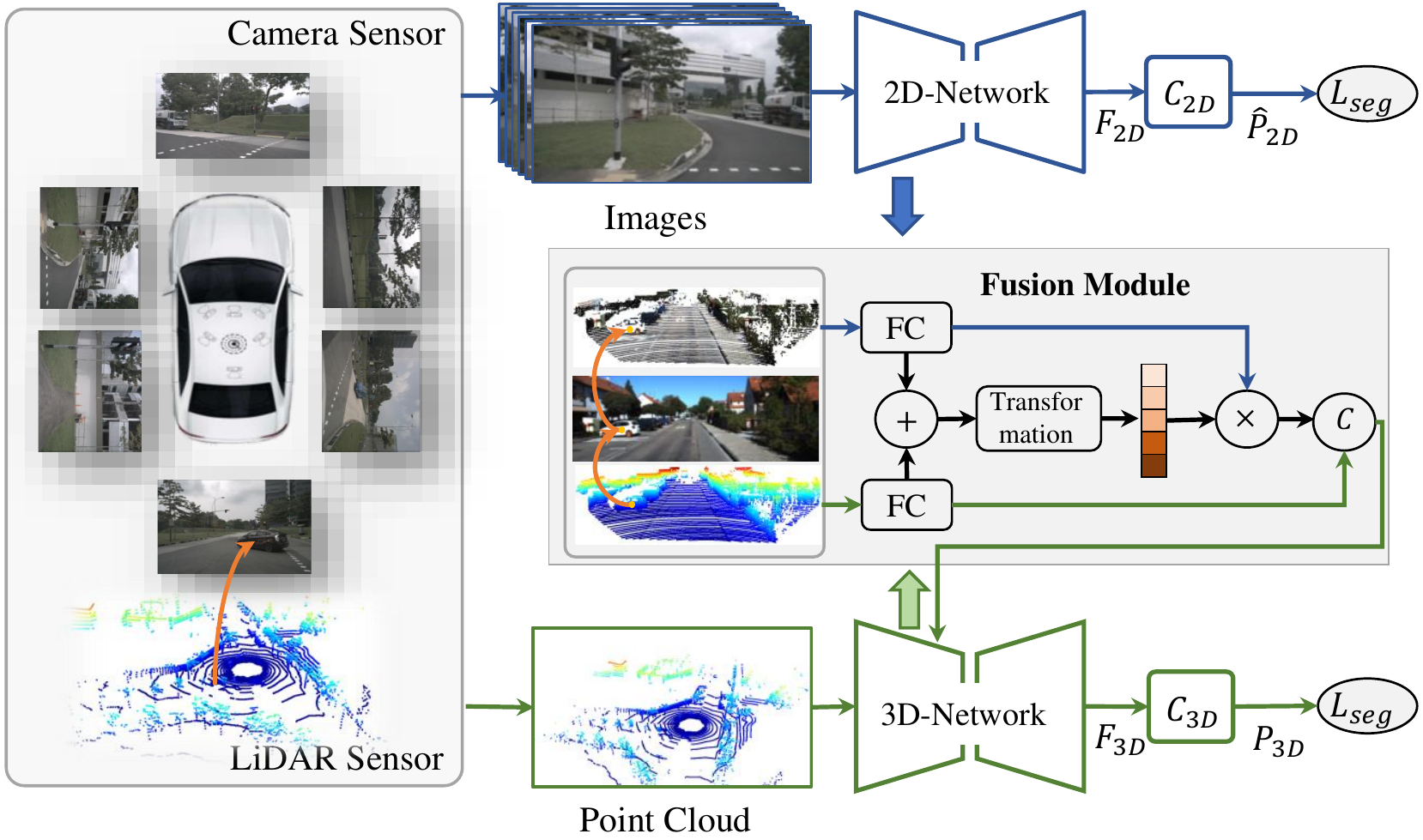}
		\caption{Graphical illustration of the framework adopted in LiDAR-RGB semantic segmentation.}\label{LiDAR baseline}
	\end{figure}
	
	\subsection{Proof to Thesis 2}\label{proof_C_2}
	As analyzed in \ref{proof_1}, MV$^2$D preserves all compositions included by $I^c_i$ through Eq. (\ref{MVD_4}). More specifically, as illustrated in Fig. \ref{Venn_MVD}, MV$^2$D would preserve $I_{yz_1z_2|z_3}$, $I_{yz_1z_3|z_2}$ and $I_{yz_1z_2z_3}$ for $z_1$, and perform the same processes for the remaining ones, {\it i.e.}, $z_2$, and $z_3$. As a result, the learned representations $\{z_1,z_2,z_3\}$ encode $I_{yz_1z_2z_3}$ with a larger weight since it is shared by every viewpoint. Correspondingly, other components ({\it e.g.}, $I_{yz_1z_2|z_3}$, $I_{yz_1z_3|z_2}$) are assigned with smaller weights due to the partial sharing. On this basis, the preserved information would be automatically prioritized by MV$^2$D, which proves ({\textit{$T_2$}}). 
	
	\section{FRAMEWORK DETAILS}
	\subsection{Cross-Modal Person Re-identification}
	To further investigate the effectiveness of the proposed MV$^2$D in 3-view circumstance, we adopt another viewpoint in addition to the default infrared and visible ones.  More specifically, we deploy the U-Net \cite{unet} trained with Eq. (\ref{reconstruction loss}) to obtain the additional viewpoint.
	\begin{equation}
		\mathcal{L}_{recon}=\omega \cdot \ell_2\left(x, x^{\prime}\right)+\operatorname{Lap}_{1}\left(x, x^{\prime}\right).\label{reconstruction loss}
	\end{equation}
	Note $\ell_{2}\left(x, x^{\prime}\right)=\left\|x-x^{\prime}\right\|_{2}^{2}$ is the squared-loss function, and $\operatorname{Lap}_{1}\left(x, x^{\prime}\right)$ denotes the Laplacian pyramid loss \cite{Laplacian} defined as:	
	\begin{equation}
		\operatorname{Lap}_{1}\left(x, x^{\prime}\right)=\sum_{j} 2^{2 j}\left|L^{j}(x)-L^{j}\left(x^{\prime}\right)\right|_{1},
		\label{eq1}
	\end{equation}
	where $L^{j}(x)$ is the $j$-th level of the Laplacian pyramid representation of $x$.

	\begin{figure}[t]
		\centering
		\includegraphics[width=1\linewidth]{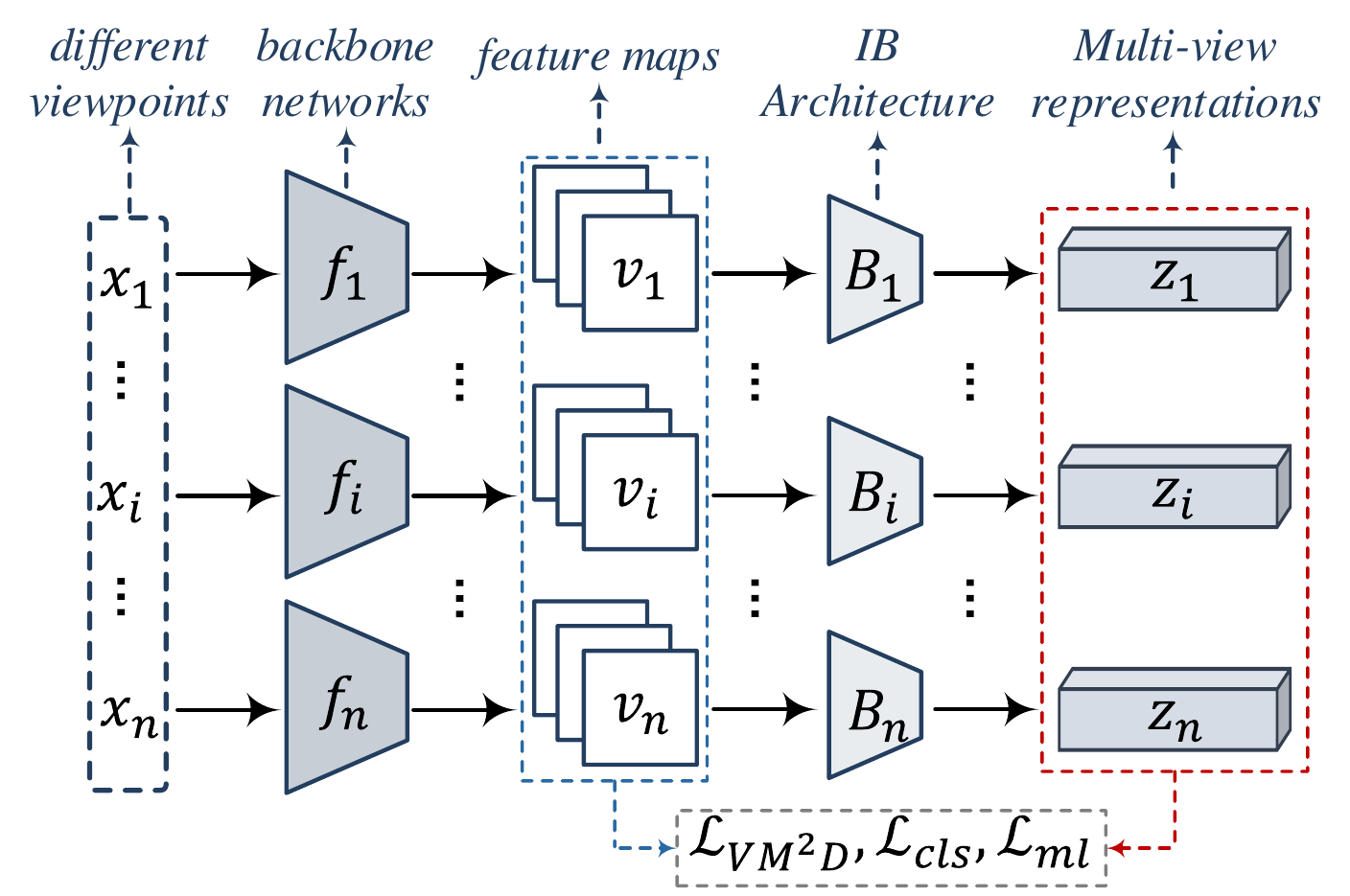}
		\caption{Overall framework utilized for multi-view classification, where the subscriptions index different viewpoint.}\label{multi-view baseline}
	\end{figure}

	\subsection{LiDAR-RGB Semantic Segmentation}
	In this section, we elaborate the framework adopted in LiDAR-RGB semantic segmentation. As is shown in Fig. {\ref{LiDAR baseline}}, we deploy two branches to handle inputs from both RGB and LiDAR sensors, and each of which is composed of one backbone network and one information bottleneck architecture to enable the use of VCD. Specifically, the 2-D network and 3-D network are implemented with SwiftNet \cite{SwiftNet} and SPVCNN \cite{SPVCNN}, respectively. Besides, we also adopt a fusion module \cite{EPNet} which applies various transformation ({\it e.g.}, element-wise addition/product, concatenation) to the inputs of image and point cloud. The fused outputs are then utilized to facilitate 3-D representation learning. Note the segmentation loss $\mathcal{L}_{seg}$ consists of multi-class focal loss \cite{focal_loss}, Lov$\acute{{\text{a}}}$sz-softmax loss \cite{Lovasz_loss} and our VCD for better performance.
	
	\subsection{Multi-View Classification}
	In this section, we show more details of the adopted framework for multi-view classification. As is illustrated in Fig. \ref{multi-view baseline}, $\{f_i\}^{n}_{i=1}$ is a set of backbone networks utilized to transform the input to feature maps, which are then encoded into a series of multi-view representation with much lower dimensions. Note both $\{v_i\}^{n}_{i=1}$ and $\{z_i\}^{n}_{i=1}$ are entailed for loss computation, with $\mathcal{L}_{MV^2D}$, $\mathcal{L}_{cls}$ and $\mathcal{L}_{ml}$ denoted as the proposed multi-view variational distillation, cross-entropy and metric learning.

~\\






\bibliographystyle{IEEEtran}
\bibliography{egbib}




\clearpage

\ifCLASSOPTIONcaptionsoff
\newpage
\fi



%

%

\end{document}